%% file: igwpap.tex
\documentclass[11pt]{article}
\usepackage{graphicx,subfigure,ifthen,color,algorithm,palatino}
\usepackage{amsmath,amsthm,amssymb,amsfonts,verbatim}
\usepackage{mathrsfs,accents,setspace,booktabs}
\include{igwmacros}
\newboolean{UnBlinded}
\newboolean{DoubleSpaced}
\setboolean{UnBlinded}{true}
\setboolean{DoubleSpaced}{false}
\newtheorem{definition}{\textbf{Definition}}
\newtheorem{result}{\textbf{Result}}
\begin{document}

\ifthenelse{\boolean{DoubleSpaced}}{\setstretch{1.5}}{}

\vskip5mm
\centerline{\Large\bf The Inverse G-Wishart Distribution and Variational Message Passing}
\vskip5mm
\centerline{\normalsize\sc By L. Maestrini and M.P. Wand}
\vskip5mm
\centerline{\textit{University of Technology Sydney}}
\vskip6mm
\centerline{10th December, 2020}

\vskip6mm

\centerline{\large\bf Abstract}
\vskip2mm

Message passing on a factor graph is a powerful paradigm for the coding
of approximate inference algorithms for arbitrarily large graphical models. 
The notion of a factor graph fragment allows for compartmentalization of algebra and
computer code. We show that the Inverse G-Wishart family of distributions 
enables fundamental variational message passing factor graph fragments 
to be expressed elegantly and succinctly. Such fragments arise in  
models for which approximate inference concerning covariance matrix
or variance parameters is made, and are ubiquitous in contemporary
statistics and machine learning.


\section{Introduction}\label{sec:intro}

We argue that a very general family of covariance matrix distributions, known
as the \emph{Inverse G-Wishart} family, plays a fundamental role in modularization
of variational inference algorithms via variational message passing when a factor 
graph fragment (Wand, 2017) approach is used. A factor graph fragment, or 
\emph{fragment} for short, is a sub-graph of the relevant factor graph 
consisting of a factor and all of its neighboring nodes. Even though use of the 
Inverse G-Wishart distribution is not necessary, its adoption allows for fundamental 
factor graph fragment natural parameter updates to be expressed elegantly and succinctly.
An essential aspect of this strategy is that the Inverse G-Wishart distribution
is the \emph{only} distribution used for covariance matrix and variance parameters.
The family includes as special cases the Inverse Chi-Squared, Inverse Gamma and 
Inverse Wishart distributions. Therefore, just a single distribution is required
which leads to savings in notation and code. Whilst similar comments concerning modularity
apply to Monte Carlo-based approaches to approximate Bayesian inference, here
we focus on variational inference.

 Two of the most common contemporary approaches to fast approximate Bayesian inference
are mean field variational Bayes (e.g.\ Attias, 1999) and expectation propagation
(e.g.\ Minka, 2001). Minka (2005) explains how each approach can be expressed 
as message passing on relevant \emph{factor graphs} with \emph{variational message passing}
(Winn \myand Bishop, 2005) being the name used for the message passing version
of mean field variational Bayes. Wand (2017) introduced the concept of \emph{factor graph
fragments}, or \emph{fragments} for short, for compartmentalization of variational
message passing into atom-like components. Chen \myand Wand (2020) demonstrate
the use of fragments for expectation propagation. Explanations of factor graph-based
variational message passing that match the current exposition are given in 
Sections 2.4--2.5 of Wand (2017).

  Sections 4.1.2--4.1.3 of Wand (2017) introduce two variational message passing fragments
known as the \emph{Inverse Wishart prior} fragment and the \emph{iterated Inverse G-Wishart}
fragment. The first of these simply corresponds to imposing an Inverse Wishart prior
on a covariance matrix. In the scalar case this reduces to imposing an Inverse Chi-Squared
or, equivalently, an Inverse Gamma prior on a variance parameter. The iterated Inverse G-Wishart 
fragment facilitates the imposition of arbitrarily non-informative priors on 
standard deviation parameters such as members of the Half-$t$ family (Gelman, 2006
; Polson \myand Scott, 2012). An extension to the covariance matrix case, 
for which there is the option to impose marginal Uniform distribution priors 
over the interval $(-1,1)$ on correlation parameters, 
is elucidated in Huang \myand Wand (2013). Mulder \myand Pericchi (2018) 
provide a different type of extension that is labelled the \emph{Matrix-$F$}
distribution. These two fragments arise in many classes of Bayesian models, 
such as both Gaussian and generalized response linear mixed models 
(e.g. McCulloch \textit{et al.}, 2008), Bayesian factor models 
(e.g.  Conti \textit{et al.}, 2014), vector autoregressive models 
(e.g. Assaf \textit{et al.}, 2019), and generalized additive mixed models 
and group-specific curve models (e.g. Harezlak \textit{et al.}, 2018).

Despite the fundamentalness of Inverse G-Wishart-based fragments for variational 
message passing, the main reference to date, Wand (2017), is brief in its exposition
and contains some errors that affect certain cases. In this article we provide
a detailed exposition of the Inverse G-Wishart distribution in the context of 
variational message passing and list the Inverse Wishart prior and iterated Inverse G-Wishart
fragment updates in full ready-to-code forms. \textsf{R} functions 
(\textsf{R} Core Team, 2020) that implement these algorithms are provided as 
part of the supplementary material of this article. 
We also explain the errors in Wand (2017).

Section \ref{sec:GWandIGW} contains relevant definitions and results concerning
the G-Wishart and Inverse G-Wishart distributions. Connections with the
Huang-Wand and Matrix-$F$ families of marginally noninformative prior distributions for
covariance matrices are summarized in Section \ref{sec:HWandMatrixFconns} and
in Section \ref{sec:VMPbackground} we point to background material
on variational message passing. In Sections \ref{sec:IGWprior} 
and \ref{sec:iterIGWfrag} we provide detailed accounts 
of the two variational message passing fragments pertaining 
to variance and covariance matrix parameters, expanding on what is presented 
in Sections 4.1.2 and 4.1.3 of Wand (2017), and making some corrections to what 
is presented there. In Section \ref{sec:priorSpec} we provide explicit instructions on
how the two fragments are used to specify different types of prior distributions 
on standard deviation and covariance matrix parameters in variational message 
passing-based approximate Bayesian inference. Section \ref{sec:illustrative} 
contains a data analytic example that illustrates the use of the covariance matrix 
fragment update algorithms. Some closing discussion is given in Section \ref{sec:closing}.
A web-supplement contains relevant details.

\section{The G-Wishart and Inverse G-Wishart Distributions}\label{sec:GWandIGW}

A random matrix $\bX$ has an Inverse G-Wishart distribution if and only if 
$\bX^{-1}$ has a G-Wishart distribution. In this section we first review
the G-Wishart distribution, which has an established literature.
Then we discuss the Inverse G-Wishart distribution and list properties
that are relevant to its employment in variational message passing.

\begin{figure}[h]
\null\vskip4mm
$\Mone\quad\Mtwo\quad\Mthree\quad\Mfour$
\centering
     \subfigure{\includegraphics[width=0.22\textwidth]{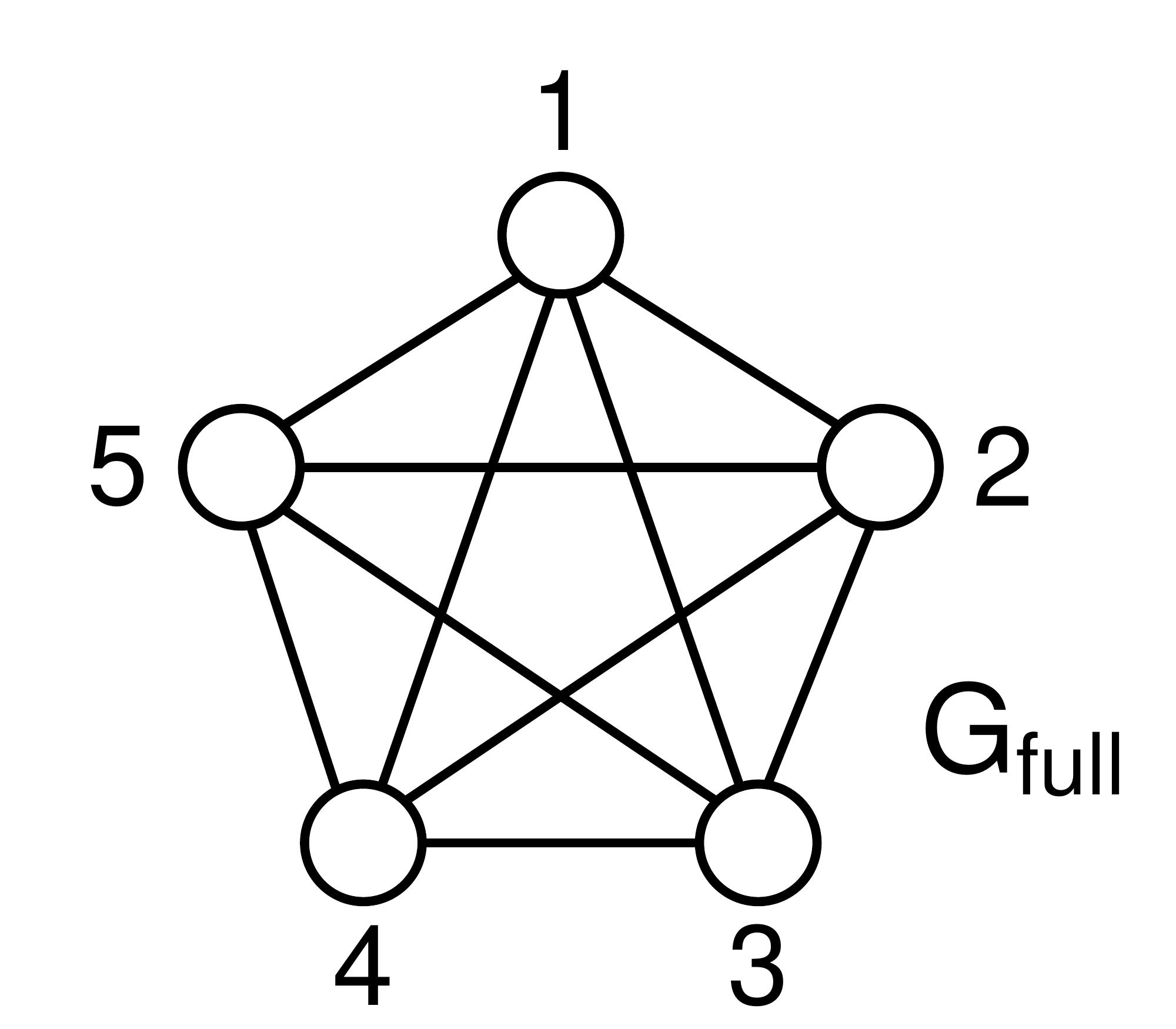}}
\quad\subfigure{\includegraphics[width=0.22\textwidth]{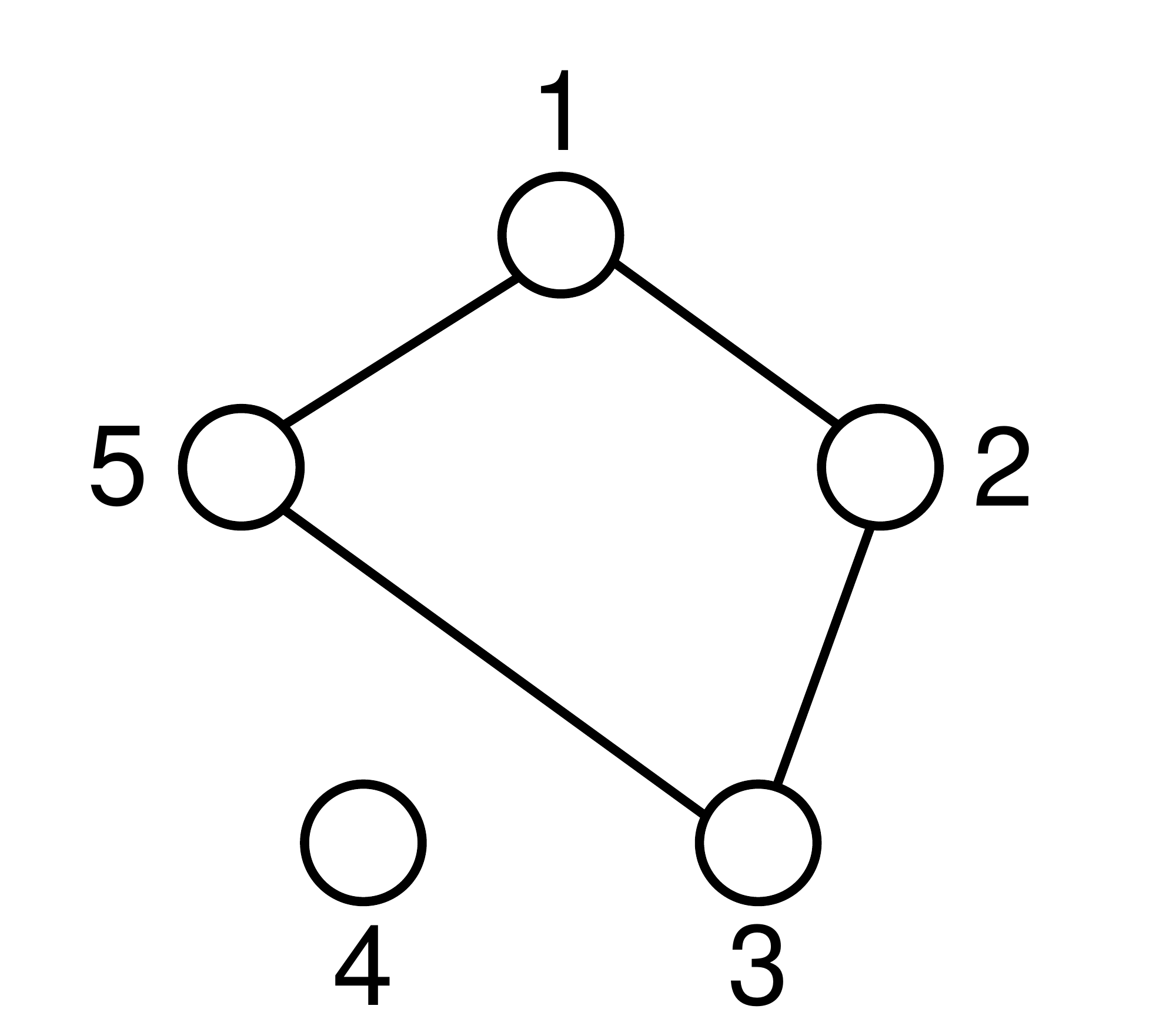}}
\quad\subfigure{\includegraphics[width=0.22\textwidth]{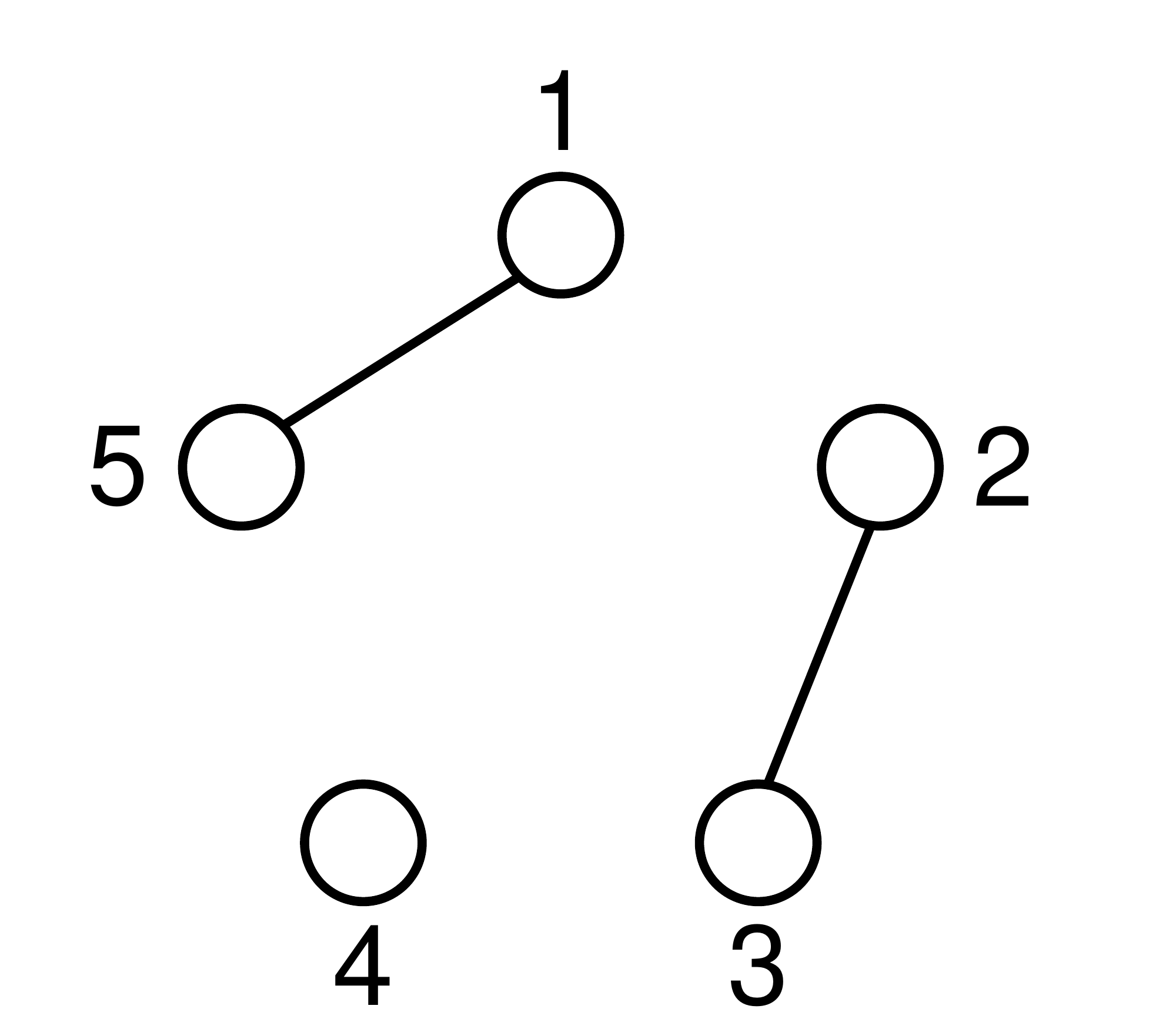}}
\quad\subfigure{\includegraphics[width=0.22\textwidth]{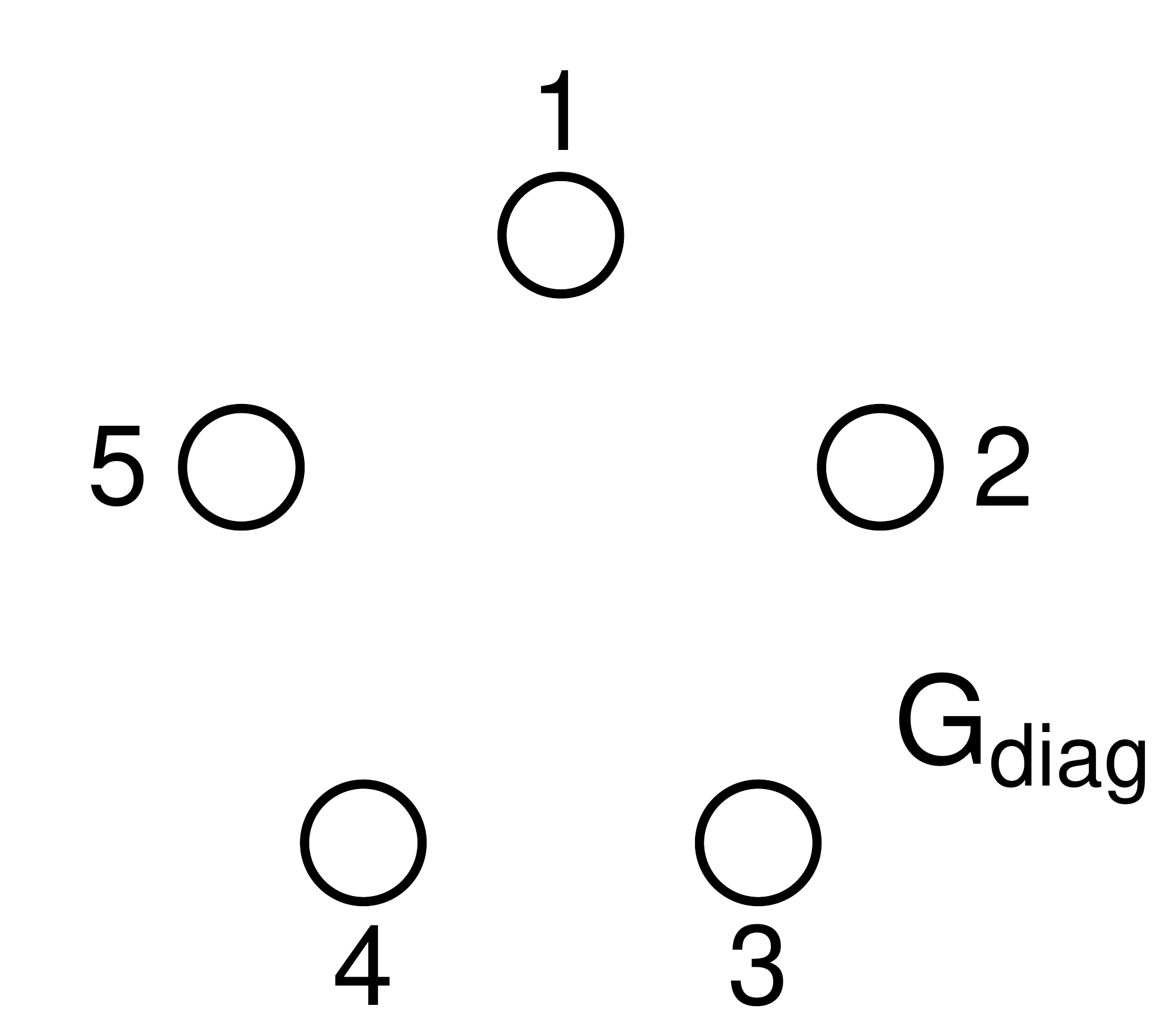}}
\caption{\textit{The zero/non-zero entries of four $5\times5$ symmetric matrices
with non-zero entries denoted by $\bigtimes$. Underneath each matrix is
the $5$-node undirected graph that the matrix respects.
The nodes are numbered according to the rows and
columns of the matrices. A graph edge is present between nodes $i$ and $j$ whenever
the $(i,j)$ entry of the matrix is non-zero. The graph respected by the
full matrix is denoted by} $\Gfull$. \textit{The graph respected by the 
diagonal matrix is denoted by} $\Gdiag$.}
\label{fig:MandG}
\end{figure}

Let $G$ be an undirected graph with $d$ nodes labeled $1,\ldots,d$ and
set $E$ consisting of pairs of nodes that are connected by an edge. 
We say that the symmetric $d\times d$ matrix $\bM$ 
\emph{respects} $G$ if 
$$\bM_{ij}=0\quad\mbox{for all}\quad \{i,j\}\notin E.$$
Figure \ref{fig:MandG} shows the zero/non-zero entries of 
four $5\times5$ symmetric matrices. For each matrix, the $5$-node graph
that the matrix respects is shown underneath.


The first graph in Figure \ref{fig:MandG} is totally connected and
corresponds to the matrix being full. Hence we denote this graph
by $\Gfull$. At the other end of the spectrum is the last graph of
Figure  \ref{fig:MandG}, which is totally disconnected. Since this
corresponds to the matrix being diagonal we denote this graph by 
$\Gdiag$. 

An important concept in G-Wishart and Inverse G-Wishart distribution theory 
is graph decomposability. An undirected graph $G$ is \emph{decomposable} if
and only if all cycles of four or more nodes have an edge that is 
not part of the cycle but connects two nodes 
of the cycle. In Figure \ref{fig:MandG} the first, third and fourth graphs are
decomposable. However, the second graph is not decomposable since it 
contains a four-node cycle that is devoid of edges that connect pairs
of nodes within this cycle. Alternative labels 
for decomposable graphs are \emph{chordal} graphs and \emph{triangulated} graphs.

In Sections \ref{sec:GWishartDistr} and \ref{sec:IGWdistn}
we define the G-Wishart and Inverse G-Wishart distributions and
treat important special cases. This exposition depends on particular
notation, which we define here. For a generic proposition $\Psc$ we
define $I(\Psc)$ to equal $1$ if $\Psc$ is true and zero otherwise.
If the random variables $x_j$, $1\le j\le d$, are independent
such that $x_j$ has distribution $\Dsc_j$ we write $x_j\simind\Dsc_j$,
$1\le j\le d$. For a $d\times 1$ vector $\bv$ let $\diag(\bv)$ be
the $d\times d$ diagonal matrix with diagonal comprising the entries of $\bv$
in order. For a $d\times d$ matrix $\bM$ let $\mbox{diagonal}(\bM)$ denote the 
$d\times 1$ vector comprising the diagonal entries of $\bM$ in order.
The $\vecof$ and $\vech$ matrix operators are well-established
(e.g. Gentle, 2007). If $\ba$ is a $d^2\times1$ vector then $\vecof^{-1}(\ba)$ is the
$d\times d$ matrix such that $\vecof\big(\vecof^{-1}(\ba)\big)=\ba$.
The matrix $\bD_d$, known as the \emph{duplication matrix of order $d$}, 
is the $d^2\times\{\frac{1}{2}d(d+1)\}$ matrix containing only zeros and ones 
such that $\bD_d\vech(\bA)=\vecof(\bA)$ for any symmetric $d\times d$ matrix $\bA$ 
(Magnus \myand Neudecker, 1999). For example,
$$\bD_2=\left[
\begin{array}{ccc}
1 & 0 & 0 \\
0 & 1 & 0 \\
0 & 1 & 0 \\
0 & 0 & 1 
\end{array}
\right].
$$
The Moore-Penrose inverse of $\bD_d$ 
is $\bD_d^+\equiv(\bD_d^T\bD_d)^{-1}\bD_d^T$ and is such that
$\bD_d^+\vecof(\bA)=\vech(\bA)$ for a symmetric matrix $\bA$.

\subsection{The G-Wishart Distribution}\label{sec:GWishartDistr}

The \textit{G-Wishart distribution} (Atay-Kayis \myand Massam, 2005) 
is defined as follows:
%
\begin{definition}
Let $\bX$ be a $d\times d$ symmetric and 
	positive definite random matrix and $G$ be a $d$-node undirected graph 
	such that $\bX$ respects $G$. For $\delta>0$ and a symmetric
        positive definite $d\times d$ matrix $\bLambda$ we say that 
        $\bX$ has a G-Wishart distribution with graph $G$, shape parameter
        $\delta$ and rate matrix $\bLambda$, and write
$$\bX\sim\mbox{\rm G-Wishart}(G,\delta,\bLambda),$$
if and only if the non-zero values of the density function of $\bX$ satisfy
\begin{equation}
\pDens(\bX)\propto|\bX|^{(\delta-2)/2}\exp\{-\smhalf\mbox{\rm tr}(\bLambda\bX)\}.
\label{eq:GWishartDensKern}
\end{equation}
\label{def:GWdefn}
\end{definition}

Obtaining an expression for the normalizing factor of a general G-Wishart density function is a 
challenging problem and recently was resolved by Uhler \textit{et al.} (2018). In the special case
where $G$ is a decomposable graph a relatively simple expression for the normalizing
factor exists and is given, for example, by equation (1.4) of Uhler \textit{et al.} (2018).
The non-decomposable case is much more difficult and treated in Section 3 of 
Uhler \textit{et al.} (2018), but the normalizing factor does not have a succinct expression
for general $G$. Similar comments apply to expressions for the mean of a G-Wishart random matrix.
As discussed in Section 3 of Atay-Kayis \myand Massam (2005), the G-Wishart distribution has
connections with other distributional constructs such as the hyper Wishart law defined by 
Dawid \myand Lauritzen (1993).

Let $\Gfull$ be the totally connected $d$-node undirected graph and $\Gdiag$ be 
the totally disconnected $d$-node undirected graph. The special cases of $G=\Gfull$ 
and $G=\Gdiag$ are such that the normalizing factor and mean do have simple closed 
form expressions. Since these cases arise in fundamental variational message passing 
algorithms we now turn our attention to them.

\subsubsection{The $G=\Gfull$ Special Case}

In the case where $G$ is a fully connected graph we have:

\begin{result}
If the $d\times d$ random matrix $\bX$ is such that 
$\bX\sim\mbox{\rm G-Wishart}(\Gfull,\delta,\bLambda)$ then
\begin{equation}
\begin{array}{rcl}
	\pDens(\bX)&=&\frac{|\bLambda|^{(\delta+d-1)/2}}{2^{d(\delta+d-1)/2}\pi^{d(d-1)/4}
		\prod_{j=1}^d\Gamma(\frac{\delta+d-j}{2})}\,
	|\bX|^{(\delta-2)/2}\exp\{-\smhalf\mbox{\rm tr}(\bLambda\bX)\}\\[1ex]
	&&\quad\times 
	I(\bX\ \mbox{a symmetric and positive definite $d\times d$ matrix}).
\end{array}
\label{eq:pXfullCase}
\end{equation}
The mean of $\bX$ is
$$E(\bX)=(\delta+d-1)\,\bLambda^{-1}.$$
\label{res:GfullResult}
\end{result}

Result \ref{res:GfullResult} is not novel at all since the $G=\Gfull$ 
case corresponds to $\bX$ having a Wishart distribution. 
In other words, (\ref{eq:pXfullCase}) is simply the density function of
a Wishart random matrix. However, it is worth pointing out the 
the shape parameter used here is different from that commonly used
for the Wishart distribution. For example, in Table A.1 of
Gelman \textit{et al.} (2014) the shape parameter is denoted by $\nu$ and
is related to the shape parameter of (\ref{eq:pXfullCase}) according to
$$\nu=\delta+d-1$$
and therefore are the same only in the special case of $\bX$ being scalar.
Also, note that Definition \ref{def:GWdefn} and  Result \ref{res:GfullResult} 
use the rate matrix parameterisation, whereas Table A.1 of
Gelman \textit{et al.} (2014) uses the scale matrix parameterisation
for the Wishart distribution. The scale matrix is $\bLambda^{-1}$.

\subsubsection{The $G=\Gdiag$ Special Case}

Before treating the $\bX\sim\mbox{\rm G-Wishart}(\Gdiag,\delta,\bLambda)$ situation,
we define the notation
\begin{equation}
x\sim\mbox{Gamma}(\alpha,\beta)
\label{eq:FriedCigar}
\end{equation}
to mean that the scalar random variable $x$ has a Gamma distribution with 
shape parameter $\alpha$ and rate parameter $\beta$. The density function corresponding
to (\ref{eq:FriedCigar}) is
$$\pDens(x)=\frac{\beta^{\alpha}}{\Gamma(\alpha)}\,x^{\alpha-1}\exp(-\beta\,x)I(x>0).$$

The $\mbox{\rm G-Wishart}(\Gdiag,\delta,\bLambda)$ distribution is tied intimately
to the Gamma distribution, as Result \ref{res:GdiagResult} shows.

%
\begin{result} 
Suppose that the $d\times d$ random matrix $\bX$ is such that 
$\bX\sim\mbox{\rm G-Wishart}(\Gdiag,\delta,\bLambda)$.
Then the non-zero entries of $\bX$ satisfy
$$X_{jj}\simind\mbox{Gamma}\big(\smhalf\delta,\smhalf\Lambda_{jj}\big),\quad 1\le j\le d,$$
where $\Lambda_{jj}$ is the $j$th diagonal entry of $\bLambda$.
The density function of $\bX$ is
{\setlength\arraycolsep{3pt}
\begin{eqnarray*}
\pDens(\bX)&=&\frac{|\bLambda|^{\delta/2}}{2^{d\delta/2}\Gamma(\delta/2)^d}
\,|\bX|^{(\delta-2)/2}\exp\{-\smhalf\mbox{\rm tr}(\bLambda\bX)\}\prod_{j=1}^d I(X_{jj}>0)\\[1ex]
&=&\frac{\prod_{j=1}^d\Lambda_{jj}^{\delta/2}}{2^{d\delta/2}\Gamma(\delta/2)^d}\,
\prod_{j=1}^d\,X_{jj}^{(\delta-2)/2}\,\exp\left(-\smhalf\sum_{j=1}^d\Lambda_{jj}\,X_{jj}\right)
\prod_{j=1}^d I(X_{jj}>0).
\end{eqnarray*}
}
The mean of $\bX$ is
$$E(\bX)=\delta\,\bLambda^{-1}=\delta\,\mbox{\rm diag}(1/\Lambda_{11},\ldots,1/\Lambda_{dd}).$$
\label{res:GdiagResult}
\end{result}

\noindent
We now make some remarks concerning Result \ref{res:GdiagResult}.
\begin{enumerate}
\item When $G=\Gdiag$ the off-diagonal entries of $\bLambda$ have no effect on
the distribution of $\bX$. In other words, the declaration 
$\bX\sim\mbox{\rm G-Wishart}(\Gdiag,\delta,\bLambda)$ is equivalent to the declaration
$\bX\sim\mbox{\rm G-Wishart}\big(\Gdiag,\delta,\diag\{\diagonal(\bLambda)\}\big)$.
\item The declaration $\bX\sim\mbox{\rm G-Wishart}(\Gdiag,\delta,\bLambda)$ is equivalent
to the diagonal entries of $\bX$ being independent Gamma random variables with shape parameter 
$\smhalf\delta$ and rate parameters equalling the diagonal entries of $\smhalf\bLambda$.
\item Even though statements concerning the distributions of independent random variables
may seem simpler than a statement of the form $\bX\sim\mbox{\rm G-Wishart}(\Gdiag,\delta,\bLambda)$,
the major thrust of this article is the elegance provided by key variational message passing
fragment updates being expressed in terms of a single family of distributions.
\end{enumerate}

\subsubsection{Exponential Family Form and Natural Parameterisation}

Suppose that $\bX\sim\mbox{\rm G-Wishart}(G,\delta,\bLambda)$. Then for $\bX$ such that $\pDens(\bX)>0$
we have
\begin{equation}
\pDens(\bX)\propto \exp\left\{
	\left[
	\begin{array}{c}
		\log|\bX|\\
		\vech(\bX)
	\end{array}
	\right]^T
	\left[
	\begin{array}{c}
		\smhalf(\delta-2)\\
		-\smhalf\bD_d^T\vecof(\bLambda)
	\end{array}
	\right]
	\right\}\\[1ex]=\exp\{\bT(\bX)^T\bdeta\}
\label{eq:vechFirst}
\end{equation}
where
$$
\bT(\bX)\equiv\left[
\begin{array}{c}
\log|\bX|\\
\vech(\bX)
\end{array}
\right]
\quad\mbox{and}\quad
\bdeta
\equiv
\left[
\begin{array}{c}
\eta_1\\
\bdeta_2
\end{array}
\right]
=
\left[
\begin{array}{c}
\smhalf(\delta-2)\\
-\smhalf\bD_d^T\vecof(\bLambda)
\end{array}
\right]
$$
are, respectively, sufficient statistic and natural parameter vectors.  The inverse of the natural parameter mapping is
$$
\left\{
\setlength\arraycolsep{1pt}{
\begin{array}{rcl}
\delta&=&2(\eta_1+1),\\[1ex]
\bLambda&=&-2\,\vecof^{-1}(\bD_d^{+T}\bdeta_2)
\end{array}.
}
\right.
$$
Note that, throughout this article, we use  
$\vech(\bX)$ rather than $\vecof(\bX)$ since the 
former is more compact and avoids duplications.
Section \ref{sec:vecANDvech} in the web-supplement 
has further discussion on this matter.

\subsection{The Inverse G-Wishart Distribution}\label{sec:IGWdistn}

Suppose that $\bX\sim\mbox{\rm G-Wishart}(G,\delta,\bLambda)$,
where $\bX$ is $d\times d$, and $\bY=\bX^{-1}$.
Let the density functions of $\bX$ and $\bY$ be denoted by $\pDens_{\bX}$ 
and $\pDens_{\bY}$  respectively. 
Then the density function of $\bY$ is
\begin{equation}
\pDens_{\bY}(\bY)=\pDens_{\bX}(\bY^{-1})\,|J(\bY)|
\label{eq:IGWdensFunction}
\end{equation}
where
$$J(\bY)\equiv\mbox{the determinant of}\ \frac{\partial\vecof(\bY^{-1})}{\partial\vecof(\bY)^T}$$
is the Jacobian of the transformation. 

An important observation is that the form of $J(\bY)$ 
is dependent on the graph $G$. In the case of $G$ being a 
decomposable graph an expression for $J(\bY)$ is given by
(2.4) of Letac \myand Massam (2007), with credit given to Roverato (2000).
Therefore, if $G$ is decomposable, the density 
function of an Inverse G-Wishart random matrix can be obtained by substitution 
of (2.4) of Letac \myand Massam (2007) into (\ref{eq:IGWdensFunction}). 
However, depending on the complexity of $G$, simplification
of the density function expression may be challenging.

With variational message passing in mind, we now turn to the $G=\Gfull$ and $G=\Gdiag$ special
cases. The $G=\Gdiag$ case is simple since it involves products of univariate density functions
and we have
\begin{equation}
\mbox{if $G=\Gdiag$ then $|J(\bY)|=|\bY|^{-2}$ for any $d\in\naturalNumbers$}.
\label{eq:JacForDiag}
\end{equation}
The $G=\Gfull$ case is more challenging and is the focus of 
Theorem 2.1.8 of Muirhead (1982):
\begin{equation}
\mbox{if $G=\Gfull$ then $|J(\bY)|=|\bY|^{-(d+1)}$}.
\label{eq:JacForFull}
\end{equation}
This result is also stated as Lemma 2.1 in Letac \myand Massam (2007).

Combining (\ref{eq:IGWdensFunction}), (\ref{eq:JacForDiag}) and (\ref{eq:JacForFull})
we have:

\begin{result}
Suppose that $\bY=\bX^{-1}$ where $\bX\sim\mbox{\rm G-Wishart}(G,\delta,\bLambda)$
and $\bX$ is $d\times d$. 
\begin{itemize}
	\item[(a)] If $G=\Gfull$ then $\pDens(\bY)\propto|\bY|^{-(\delta+2d)/2}
                   \exp\{-\smhalf\mbox{\rm tr}(\bLambda\bY^{-1})\}$.
	\item[(b)] If $G=\Gdiag$ then $\pDens(\bY)\propto|\bY|^{-(\delta+2)/2}
                   \exp\{-\smhalf\mbox{\rm tr}(\bLambda\bY^{-1})\}$.
\end{itemize}
\label{res:GWtoIGW}
\end{result}

Whilst Result \ref{res:GWtoIGW} only covers $G=\Gfull$ or $G=\Gdiag$ it shows that,
in these special cases, the density function of an Inverse G-Wishart random matrix $\bY$
is proportional to a power of $|\bY|$ multiplied by an an exponentiated trace of a matrix
multiplied by $\bY^{-1}$. This form does not necessarily arise for $G\notin\{\Gfull,\Gdiag\}$.
Since the motivating variational message passing fragment update algorithms only involve the
$G\in\{\Gfull,\Gdiag\}$ cases we focus on them for the remainder of this section.

\subsubsection{The Inverse G-Wishart Distribution When $G\in\{\Gfull,\Gdiag\}$}

For succinct statement of variational message passing fragment update algorithms
involving variance and covariance matrix parameters it is advantageous to have
a single Inverse G-Wishart distribution notation for the $G\in\{\Gfull,\Gdiag\}$ 
cases.

\begin{definition}
Let $\bX$ be a $d\times d$ symmetric and 
	positive definite random matrix and $G$ be a $d$-node undirected graph 
	such that $\bX^{-1}$ respects $G$. Let $\xi>0$ and $\bLambda$ be a symmetric
        positive definite $d\times d$ matrix $\bLambda$.
\begin{itemize}
	\item[(a)] If $G=\Gfull$ and $\xi$ is restricted such that $\xi>2d-2$ then we
                   say that $\bX$ has an Inverse G-Wishart distribution with graph $G$,
                   shape parameter $\xi$ and scale matrix $\bLambda$, and write
                   $$\bX\sim\mbox{\rm Inverse G-Wishart}(G,\xi,\bLambda),$$
                   if and only if the non-zero values of the density function of $\bX$ satisfy
$$\pDens(\bX)\propto|\bX|^{-(\xi+2)/2}\exp\{-\smhalf\mbox{\rm tr}(\bLambda\bX^{-1})\}.$$
	\item[(b)]If $G=\Gdiag$ then say that $\bX$ has an Inverse G-Wishart distribution with graph $G$,
                   shape parameter $\xi$ and scale matrix $\bLambda$, and write
                   $$\bX\sim\mbox{\rm Inverse G-Wishart}(G,\xi,\bLambda),$$
                   if and only if the non-zero values of the density function of $\bX$ satisfy
$$\pDens(\bX)\propto|\bX|^{-(\xi+2)/2}\exp\{-\smhalf\mbox{\rm tr}(\bLambda\bX^{-1})\}.$$
	\item[(c)] If $G\notin\{\Gfull,\Gdiag\}$ then $\bX\sim\mbox{\rm Inverse G-Wishart}(G,\xi,\bLambda)$ is not defined.
\end{itemize}
\label{def:IGWdefn}
\end{definition}

The shape parameter $\xi$ used in Definition \ref{def:IGWdefn} is a reasonable compromise between
various competing parameterisation choices for the Inverse G-Wishart distribution 
for $G\in\{\Gfull,\Gdiag\}$ and for use in variational message passing algorithms. It has
the following attractions:
\begin{itemize}
\item The exponent of the determinant in the density function expression is $-(\xi+2)/2$ regardless
of whether $G=\Gfull$ or $G=\Gdiag$, which is consistent with the G-Wishart distributional 
notation used in Definition \ref{def:GWdefn}.
\item In the $d=1$ case $\xi$ matches the shape parameter in the most common parameterisation
of the Inverse Chi-Squared distribution such as that used in Table A.1 of Gelman \textit{et al.} (2014).
\end{itemize}

In case where $\bX\sim\mbox{\rm Inverse G-Wishart}(\Gfull,\xi,\bLambda)$ we have 
the following:

\begin{result}
If the $d\times d$ random matrix $\bX$ is such that 
$\bX\sim\mbox{\rm Inverse G-Wishart}(\Gfull,\xi,\bLambda)$ then
$$
\begin{array}{rcl}
	\pDens(\bX)&=&\displaystyle{\frac{|\bLambda|^{(\xi-d+1)/2}}
                  {2^{d(\xi-d+1)/2}\pi^{d(d-1)/4}
		\prod_{j=1}^d\Gamma(\frac{\xi-d-j}{2}+1)}}\,
	|\bX|^{-(\xi+2)/2}\exp\{-\smhalf\mbox{\rm tr}(\bLambda\bX^{-1})\}\\[1ex]
	&&\quad\times 
	I(\bX\ \mbox{a symmetric and positive definite $d\times d$ matrix}).
\end{array}
$$
The mean of $\bX^{-1}$ is
$$E(\bX^{-1})=(\xi-d+1)\,\bLambda^{-1}.$$
\label{res:IGfullWres}
\end{result}

Result \ref{res:IGfullWres} follows directly from the fact that 
$\bX\sim\mbox{\rm Inverse G-Wishart}(\Gfull,\xi,\bLambda)$ if and only if 
$\bX$ has an Inverse Wishart distribution and established
results for the density function and mean of this distribution
given in, for example, Table A.1 of Gelman \textit{et al.} (2014).

We now deal with the $G=\Gdiag$ case. 

%
\begin{definition}
Let $x$ be a random variable.
For $\delta>0$ and $\lambda>0$ 
we say that the random variable $x$ has an Inverse Chi-Squared 
distribution with shape parameter $\delta$ and rate parameter
$\lambda$, and write 
$$x\sim\mbox{{\rm Inverse}-$\chi^2$}(\delta,\lambda),$$
if and only if $1/x\sim\chi^2(\delta,\lambda)$. If $x\sim\mbox{{\rm Inverse}-$\chi^2$}(\delta,\lambda)$ 
then the density function of $x$ is
$$\pDens(x)=\frac{(\lambda/2)^{\delta/2}}{\Gamma(\delta/2)}
x^{-(\delta+2)/2}\,\exp\{-(\lambda/2)\big/x\}I(x>0).
$$
\end{definition}
%

\noindent
We are now ready to state:
%
\begin{result}
Suppose that the $d\times d$ random matrix $\bX$ is such that 
$\bX\sim\mbox{\rm Inverse-G-Wishart}(\Gdiag,\xi,\bLambda)$.
Then the non-zero entries of $\bX$ satisfy
$$X_{jj}\simind\mbox{{\rm Inverse}-$\chi^2$}(\xi,\Lambda_{jj}),\quad 1\le j\le d,$$
where $\Lambda_{jj}$ is the $j$th diagonal entry of $\bLambda$.
The density function of $\bX$ is
{\setlength\arraycolsep{3pt}
\begin{eqnarray*}
\pDens(\bX)&=&\frac{|\bLambda|^{\xi/2}}{2^{d\xi/2}\Gamma(\xi/2)^d}
\,|\bX|^{-(\xi+2)/2}\exp\{-\smhalf\mbox{\rm tr}(\bLambda\bX^{-1})\}\prod_{j=1}^d I(X_{jj}>0)\\[1ex]
&=&\frac{\prod_{j=1}^d\Lambda_{jj}^{\xi/2}}{2^{d\xi/2}\Gamma(\xi/2)^d}
\prod_{j=1}^d\,X_{jj}^{-(\xi+2)/2}\,\exp\left\{-\smhalf\sum_{j=1}^d(\Lambda_{jj}/X_{jj})\right\}
\prod_{j=1}^d I(X_{jj}>0).
\end{eqnarray*}
}
The mean of $\bX^{-1}$ is
$$E(\bX^{-1})=\xi\bLambda^{-1}=\xi\,\mbox{\rm diag}(1/\Lambda_{11},\ldots,1/\Lambda_{dd}).$$
\end{result}

\subsubsection{Natural Parameter Forms and Sufficient Statistic Expectations}

Suppose that $\bX\sim\mbox{Inverse-G-Wishart}(G,\xi,\bLambda)$ where $G\in\{\Gfull,\Gdiag\}$.
Then for $\bX$ such that $\pDens(\bX)>0$,
$$
	\pDens(\bX)\propto\exp\left\{
	\left[
	\begin{array}{c}
		\log|\bX|\\
		\vech(\bX^{-1})
	\end{array}
	\right]^T
	\left[
	\begin{array}{c}
		-(\xi+2)/2\\
		-\smhalf\bD_d^T\vecof(\bLambda)
	\end{array}
	\right]
	\right\}=\exp\{\bT(\bX)^T\bdeta\}
$$
where
\begin{equation}
\bT(\bX)\equiv\left[
\begin{array}{c}
\log|\bX|\\[1ex]
\vech(\bX^{-1})
\end{array}
\right]
\quad\mbox{and}\quad
\bdeta
\equiv
\left[
\begin{array}{c}
\eta_1\\
\bdeta_2
\end{array}
\right]
=
\left[
\begin{array}{c}
-\smhalf(\xi+2)\\[1ex]
-\smhalf\bD_d^T\vecof(\bLambda)
\end{array}
\right]
\label{eq:IsawLara}
\end{equation}
are, respectively, sufficient statistic and natural parameter vectors.
The inverse of the natural parameter mapping is
\begin{equation}
\left\{
\begin{array}{rcl}
\xi&=&-2\eta_1-2,\\[1ex]
\bLambda&=&-2\,\vecof^{-1}(\bD_d^{+T}\bdeta_2).
\end{array}
\right.
\label{eq:IGWinvMap}
\end{equation}
As explained in Section \ref{sec:vecANDvech} of the web-supplement,
alternatives to (\ref{eq:IsawLara}) are those that use $\vecof(\bX)$
instead of $\vech(\bX)$. Throughout this article we use the more
compact ``$\vech$'' form.

The following result is fundamental to succinct formulation of
updates of covariance and variance parameter fragment updates 
for variational message passing:

\begin{result}
If $\bX$ is a $d\times d$ random matrix that has an Inverse G-Wishart distribution 
with graph $G\in\{\Gfull,\Gdiag\}$ and natural parameter vector $\bdeta$. Then
$$E(\bX^{-1})=\left\{    
\begin{array}{ll}
\{\eta_1+\smhalf(d+1)\}\{\vecof^{-1}(\bD_d^{+T}\bdeta_2)\}^{-1} & \mbox{if}\ \ G=\Gfull\\[1.5ex]
(\eta_1+1)\{\vecof^{-1}(\bD_d^{+T}\bdeta_2)\}^{-1}& \mbox{if}\ \ G=\Gdiag.
\end{array}
\right.
$$
\label{eq:recipMoment}
\end{result}

\subsubsection{Relationships with the Hyper Inverse Wishart Distributions}

Throughout this article we follow the G-Wishart nomenclature 
as used by, for example, Atay-Kayis \myand Massam (2005),
Letac \myand Massam (2007) and Uhler \textit{et al.} (2018)
in our naming of the Inverse G-Wishart family.
Some earlier articles, such as Roverato (2000), use the 
term \emph{Hyper Inverse Wishart} for the same
family of distributions. The naming used here is in keeping
with the more recent literature concerning Wishart distributions
with graphical restrictions.

\section{Connections with Some Recent Covariance Matrix Distributions}\label{sec:HWandMatrixFconns}

Recently Huang \myand Wand (2013) and Mulder \myand Pericchi (2018) developed
covariance matrix distributional families that have attractions in terms
of the types of marginal prior distributions that can be imposed on 
interpretable parameters within the covariance matrix. Mulder \myand Pericchi (2018) 
referred to their proposal as the \emph{Matrix-F} family of distributions.

\subsection{The Huang-Wand Family of Distributions}

A major motivation for working with the Inverse G-Wishart distribution is the fact that
the family of marginally non-informative priors proposed in Huang \myand Wand (2013) 
can be expressed succinctly in terms of the $\mbox{Inverse-G-Wishart}(G,\xi,\bLambda)$ 
family  where $G\in\{\Gfull,\Gdiag\}$. This means that variational message fragments that
cater for Huang-Wand prior specification, as well as Inverse-Wishart prior specification, 
only require natural parameter vector manipulations within a single distributional family.

If $\bSigma$ is a $d\times d$ symmetric positive definite matrix then, for 
$\nuHW>0$ and $s_1,\ldots,s_d>0$, the specification
\begin{equation}
\begin{array}{c}
\bSigma|\bA\sim\mbox{Inverse-G-Wishart}\Big(\Gfull,\nuHW+2d-2,\bA^{-1}\Big),\\[2ex]
\bA\sim\mbox{Inverse-G-Wishart}\Big(\Gdiag,1,\big\{\nuHW\,\diag(s_1^2,\ldots,s_d^2)\big\}^{-1}\Big)
\end{array}
\label{eq:HuangWandGeneralPrior}
\end{equation}
places a distribution of the type given in Huang \myand Wand (2013) 
on $\bSigma$ with shape parameter $\nuHW$ and scale parameters
$s_1,\ldots,s_d$. 

The specification (\ref{eq:HuangWandGeneralPrior}) matches (2) of Huang \myand Wand (2013) but
with some differences in notation. Firstly, $d$ is used for matrix dimension here rather
than $p$ in Huang \myand Wand (2013). Also, the $s_j$, $1\le j\le d$, 
scale parameters are denoted by $A_j$
in  Huang \myand Wand (2013). The $a_j$ auxiliary variables in (2) of Huang \myand Wand (2013)
are related to the matrix $\bA$ via the expression $\mbox{diag}(a_1,\ldots,a_d)=2\nuHW\bA$.

As discussed in Huang \myand Wand (2013), special cases of (\ref{eq:HuangWandGeneralPrior}) 
correspond to marginally noninformative prior specification of the covariance matrix $\bSigma$
in the sense that the standard deviation parameters $\sigma_j\equiv(\bSigma)^{1/2}_{jj}$,
$1\le j\le d$,
can have Half-$t$ priors with arbitrarily large scale parameters, 
controlled by the $s_j$ values.
This is in keeping with the advice given in Gelman (2006). Moreover, correlation parameters 
$\rho_{jj'}\equiv(\bSigma)^{1/2}_{jj'}/(\sigma_j\sigma_{j'})$, for each $j\ne j'$ pair, have 
a Uniform distribution over the interval $(-1,1)$ when $\nuHW=2$.
We refer to this special case as the \emph{Huang-Wand} marginally non-informative prior 
distribution with scale parameters $s_1,\ldots,s_d$ and write
\begin{equation}
\bSigma\sim\mbox{Huang-Wand}(s_1,\ldots,s_d)
\label{eq:HWprior}
\end{equation}
as a shorthand for (\ref{eq:HuangWandGeneralPrior}) with $\nuHW=2$.

\subsection{The Matrix-$F$ Family of Distributions}

For  $\nuMP>d-1$, $\deltaMP>0$ and $\bBMP$ a $d\times d$ symmetric positive 
definite matrix Mulder \myand Pericchi (2018) defined 
a $d\times d$ random matrix $\bSigma$ to have a Matrix-$F$ distribution,
written
\begin{equation}
\bSigma\sim F(\nuMP,\deltaMP,\bBMP),
\label{eq:CockOfTheNorth}
\end{equation}
if its density function has the form
$$\pDens(\bSigma)\propto \frac{|\bSigma|^{(\nuMP-d-1)/2}\,
I(\bSigma\ \mbox{symmetric and positive definite})}
{|\bI_d+\bSigma\bBMP^{-1}|^{(\nuMP+\deltaMP+d-1)/2}}.
$$
However, standard manipulations of results given in Mulder \myand Pericchi (2018) 
show that specification (\ref{eq:CockOfTheNorth}) is equivalent to
\begin{equation}
\begin{array}{c}
\bSigma|\bA\sim\mbox{Inverse-G-Wishart}\Big(\Gfull,\deltaMP+2d-2,\bA^{-1}\Big),\\[2ex]
\bA\sim\mbox{Inverse-G-Wishart}\Big(\Gfull,\nuMP+d-1,\bBMP^{-1}\Big)
\end{array}
\label{eq:FMatrixPrior}
\end{equation}
in the notation used in the current paper. An important difference
between (\ref{eq:HuangWandGeneralPrior}) and 
(\ref{eq:FMatrixPrior}) is that the former involves $\bA$ having
an Inverse G-Wishart distribution with the restriction $G=\Gdiag$, whilst the
latter has $G=\Gfull$. Section 2.4 of Mulder \myand Pericchi (2018) 
compares the two specifications in terms of the types of prior
distributions that can be imposed on standard deviation and
correlation parameters.

\section{Variational Message Passing Background}\label{sec:VMPbackground}

The overarching goal of this article is to identify and specify 
algebraic primitives for flexible imposition of covariance matrix priors
within a variational message passing framework. In Wand (2017) these
algebraic primitives are organised into fragments. This formalism is 
also used in Nolan \myand Wand (2017), Maestrini \myand Wand (2018) and 
McLean \myand Wand (2019).

  Despite it being a central theme of this article, we will not provide a 
detailed description of variational message passing here. Instead
we refer the reader to Sections 2--4 of Wand (2017) for the relevant
variational message passing background material. 

Since the notational conventions for messages used in this section's references
are used in the remainder of this article we summarize them here.
If $f$ denotes a generic factor and $\theta$ denotes a generic stochastic
variable that is a neighbour of $f$ in the factor graph then the message
passed from $f$ to $\theta$ and the message passed from $\theta$ to $f$ 
are both functions of $\theta$ and are denoted by, respectively,
$$\mSUBfTOtheta\quad\mbox{and}\quad\mSUBthetaTOf.$$
Typically, the messages are proportional to an exponential family density function
with sufficient statistic $\bT(\theta)$, and we have
$$\mSUBfTOtheta\propto\exp\left\{\bT(\theta)^T\etaSUBfTOtheta\right\}
\quad\mbox{and}\quad
\mSUBthetaTOf\propto\exp\left\{\bT(\theta)^T\etaSUBthetaTOf\right\}
$$
where $\etaSUBfTOtheta$ and $\etaSUBthetaTOf$ are the message natural
parameter vectors. Such vectors play a central role in variational message passing
iterative algorithms. We also adopt the notation
$$\etaSUBfCONNtheta\equiv\etaSUBfTOtheta+\etaSUBthetaTOf.$$

\section{The Inverse G-Wishart Prior Fragment}\label{sec:IGWprior}

The Inverse G-Wishart prior fragment corresponds to the following prior
imposition on a $d\times d$ covariance matrix $\bTheta$:
$$\bTheta\sim\mbox{Inverse-G-Wishart}(\GsubTheta,\xiSubTheta,\LambdaSubTheta)$$
for a $d$-node undirected graph $\GsubTheta$, scalar shape parameter $\xiSubTheta$
and scale matrix $\LambdaSubTheta$. The fragment's factor is 
{\setlength\arraycolsep{3pt}
\begin{eqnarray*}
\pDens(\bTheta)&\propto&|\bTheta|^{-(\xiSubTheta+2)/2}\exp\{-\smhalf\tr(\LambdaSubTheta\bTheta^{-1})\}\\[2ex]
&&\quad\times I(\bTheta\ \mbox{is symmetric and positive definite and}\ \bTheta^{-1}\ \mbox{respects}\ \GsubTheta).
\end{eqnarray*}
}
%

\begin{figure}[h]
\centering
{\includegraphics[width=0.25\textwidth]{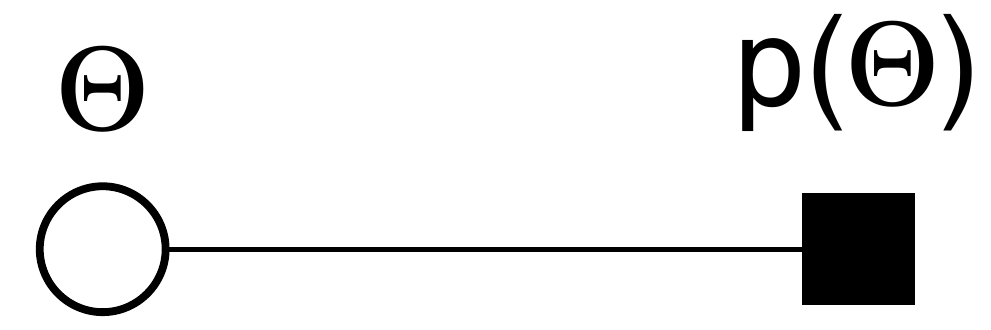}}
\caption{\it Diagram of the Inverse G-Wishart prior fragment.}
\label{fig:InvGWishPriorFrag} 
\end{figure}

Figure \ref{fig:InvGWishPriorFrag} is a diagram of the fragment, which shows that its only
factor to stochastic node message is 
$$\mSUBpThetaTOTheta\propto \pDens(\bTheta)$$
which leads to 
$$\mSUBpThetaTOTheta=\exp\left\{
\left[                
\begin{array}{c}
\log|\bTheta|\\[1ex]
\vech(\bTheta^{-1})
\end{array}
\right]^T
\left[                
\begin{array}{c}
-\smhalf\,(\xiSubTheta+2)\\[1ex]
-\smhalf\,\bD_d^T\vecof(\LambdaSubTheta)
\end{array}
\right]
\right\}.
$$
Therefore, the natural parameter update is
$$\etaSUBpThetaTOTheta\thickarrow\left[                
\begin{array}{c}
-\smhalf\,(\xiSubTheta+2)\\[1ex]
-\smhalf\,\bD_d^T\vecof(\LambdaSubTheta)
\end{array}
\right].
$$
Apart from passing the natural parameter vector out of the fragment, we
should also pass the graph out of the fragment. This entails the update:
$$\GSUBpThetaTOTheta\thickarrow \GsubTheta.$$

Algorithm \ref{alg:InvGWishPriorFrag} provides the inputs,
updates and outputs for the Inverse G-Wishart prior fragment.

\begin{algorithm}[!th]
\begin{center}
\begin{minipage}[t]{165mm}
\textbf{Hyperparameter Inputs:} $\GsubTheta,\xiSubTheta,\LambdaSubTheta$.
\jump\noindent
\textbf{Updates:}
\begin{itemize}
\setlength\itemsep{0pt}
\item[] $\etaSUBpThetaTOTheta\thickarrow\left[                
\begin{array}{c}
-\smhalf\,(\xiSubTheta+2)\\[1ex]
-\smhalf\,\bD_d^T\vecof(\LambdaSubTheta)
\end{array}
\right]$\ \ \ ;\ \ \ $\GSUBpThetaTOTheta\thickarrow \GsubTheta$
\end{itemize}
\textbf{Outputs:} $\GSUBpThetaTOTheta$, $\etaSUBpThetaTOTheta$.
\jump
\end{minipage}
\end{center}
\caption{\it The inputs, updates and outputs for the Inverse G-Wishart prior fragment.}
\label{alg:InvGWishPriorFrag} 
\end{algorithm}

\section{The Iterated Inverse G-Wishart Fragment}\label{sec:iterIGWfrag}

\begin{figure}[h]
\centering
{\includegraphics[width=0.4\textwidth]{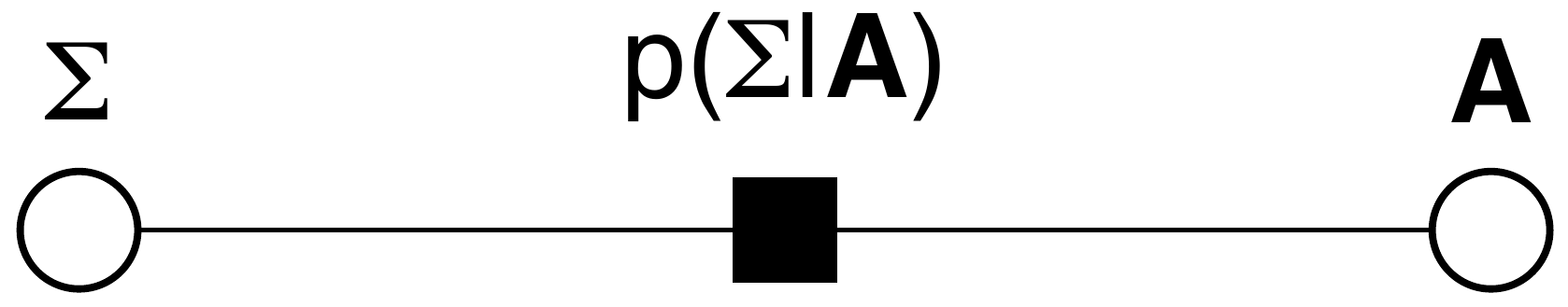}}
\caption{\it Diagram of the iterated Inverse G-Wishart fragment.}
\label{fig:iterInvGWishFrag} 
\end{figure}

The iterated Inverse G-Wishart fragment corresponds to the
following specification involving a $d\times d$ covariance
matrix $\bSigma$:
$$\bSigma|\bA\sim\mbox{Inverse-G-Wishart}(G,\xi,\bA^{-1})$$
where $G$ is a $d$-node undirected graph such that $G\in\{\Gfull,\Gdiag\}$ and
$\xi$ is a particular deterministic value of the Inverse G-Wishart shape parameter
according to Definition \ref{def:IGWdefn}. Figure \ref{fig:iterInvGWishFrag}
is a diagram of this fragment, showing that it has a
factor $\pDens(\bSigma|\bA)$ connected to two stochastic nodes
$\bSigma$ and $\bA$. 

The factor of the iterated Inverse G-Wishart fragment is,
as a function of both $\bSigma$ and $\bA$,
{\setlength\arraycolsep{3pt}
\begin{eqnarray*}
\pDens(\bSigma|\bA)\propto\left\{\begin{array}{l}
|\bA|^{-(\xi-d+1)/2}|\bSigma|^{-(\xi+2)/2}\,
\exp\{-\smhalf\tr(\bA^{-1}\bSigma^{-1})\}\quad\mbox{if $G=\Gfull$,}\\[2ex]
|\bA|^{-\xi/2}|\bSigma|^{-(\xi+2)/2}\,\exp\{-\smhalf\tr(\bA^{-1}\bSigma^{-1})\}
\quad\mbox{if $G=\Gdiag$.}
\end{array}   
\right.
\end{eqnarray*}
}
As shown in Section \ref{sec:drvFirstMsg} of the web-supplement both
of the factor to stochastic node messages of this fragment,
$$\mSUBpSigmaATOSigma\quad\mbox{and}\quad\mSUBpSigmaATOA,$$ 
are proportional to Inverse G-Wishart density functions with graph $G\in\{\Gfull,\Gdiag\}$.
We assume the following conjugacy constraints: 

\begin{center}
\begin{minipage}[t]{130mm}
All messages passed to $\bSigma$ and $\bA$ from outside 
the fragment are proportional to Inverse G-Wishart density functions 
with graph $G\in\{\Gfull,\Gdiag\}$. The Inverse G-Wishart messages
passed between $\bSigma$ and $\pDens(\bSigma|\bA)$ have the same graph.
The Inverse G-Wishart messages
passed between $\bA$ and $\pDens(\bSigma|\bA)$ have the same graph.
\end{minipage}
\end{center}

\noindent
Under these constraints, and in view of e.g. (7) of Wand (2017), the message 
passed from $\bSigma$ to $\pDens(\bSigma|\bA)$ has the form 
$$
\mSUBSigmaTOpSigmaA=
\exp\left\{ 
\left[
\begin{array}{c}
\log|\bSigma|\\[1ex]
\vech(\bSigma^{-1})
\end{array}
\right]^T\etaSUBSigmaTOpSigmaA\right\}
$$
and the message passed from $\bA$ to $\pDens(\bSigma|\bA)$ has the form 
$$
\mSUBATOpSigmaA=
\exp\left\{ 
\left[
\begin{array}{c}
\log|\bA|\\[1ex]
\vech(\bA^{-1})
\end{array}
\right]^T\etaSUBATOpSigmaA\right\}.
$$
Algorithm \ref{alg:IterInvGWishFrag} gives the full set of updates
of the message natural parameter vectors and graphs
for the iterated Inverse-G-Wishart fragment. The derivation 
of Algorithm \ref{alg:IterInvGWishFrag} is given
in Section \ref{sec:mainAlgDrv} of the web-supplement. 

\begin{algorithm}[!th]
\begin{center}
\begin{minipage}[t]{165mm}
\textbf{Graph Input:} $G\in\{\Gfull,\Gdiag\}$.\\[1ex]    
\textbf{Shape Parameter Input:} $\xi>0$.\\[1ex]    
\textbf{Message Graph Input:}\ $\GSUBATOpSigmaA\in\{\Gfull,\Gdiag\}$.\\[1ex]
\textbf{Natural Parameter Inputs:} $\etaSUBSigmaTOpSigmaA$,\ $\etaSUBpSigmaATOSigma$,\ 
$\etaSUBATOpSigmaA$,\ $\etaSUBpSigmaATOA$.\\[1ex]
\textbf{Updates:}
\begin{itemize}
\setlength\itemsep{0pt}
\item[] $\GSUBpSigmaATOSigma\thickarrow G$\ \ \ ;\ \ \ 
        $\GSUBpSigmaATOA\thickarrow\GSUBATOpSigmaA$
\item[] $\etaSUBpSigmaACONNSigma\thickarrow\etaSUBpSigmaATOSigma+\etaSUBSigmaTOpSigmaA$
\item[] $\etaSUBpSigmaACONNA\thickarrow\etaSUBpSigmaATOA+\etaSUBATOpSigmaA$
\item[] If $\GSUBpSigmaATOA=\Gfull$ then $\omega_1\thickarrow(d+1)/2$
\item[] If $\GSUBpSigmaATOA=\Gdiag$ then $\omega_1\thickarrow 1$
\item[] $E_{\qDens}(\bA^{-1})\thickarrow\Big\{\big(\etaSUBpSigmaACONNA\big)_1+\omega_1\Big\}
\left\{\vecof^{-1}\Big(\bD_d^{+T}\big(\etaSUBpSigmaACONNA\big)_2\Big)\right\}^{-1}$
\item[] If $\GSUBpSigmaATOSigma=\Gdiag$ then 
$E_{\qDens}(\bA^{-1})\thickarrow\diag\left\{\diagonal\Big(E_{\qDens}(\bA^{-1})\Big)\right\}$
\item[] $\etaSUBpSigmaATOSigma\thickarrow
\left[
\begin{array}{c}
-\smhalf\,(\xi+2)\\[1ex]
-\smhalf\bD_d^T\vecof\Big(E_{\qDens}(\bA^{-1})\Big)
\end{array}
\right]$
\item[] If $\GSUBpSigmaATOSigma=\Gfull$ then $\omega_2\thickarrow(d+1)/2$
\item[] If $\GSUBpSigmaATOSigma=\Gdiag$ then $\omega_2\thickarrow 1$
\item[] $E_{\qDens}(\bSigma^{-1})\thickarrow\Big\{\big(\etaSUBpSigmaACONNSigma\big)_1+\omega_2\Big\}
\left\{\vecof^{-1}\Big(\bD_d^{+T}\big(\etaSUBpSigmaACONNSigma\big)_2\Big)\right\}^{-1}$
\item[]If $\GSUBpSigmaATOA=\Gdiag$ then 
$E_{\qDens}(\bSigma^{-1})\thickarrow\diag\left\{\diagonal\Big(E_{\qDens}(\bSigma^{-1})\Big)\right\}$
\item[] $\etaSUBpSigmaATOA\thickarrow
\left[
\begin{array}{c}
-(\xi+2-2\omega_2)/2 \\[1ex]
-\smhalf\bD_d^T\vecof\Big(E_{\qDens}(\bSigma^{-1})\Big)
\end{array}
\right]$
\end{itemize}
\textbf{Outputs:} $\GSUBpSigmaATOSigma,\GSUBpSigmaATOA,\etaSUBpSigmaATOSigma,\etaSUBpSigmaATOA$.
\jump
\end{minipage}
\end{center}
\caption{\it The inputs, updates and outputs for the iterated Inverse G-Wishart fragment.}
\label{alg:IterInvGWishFrag} 
\end{algorithm}
\null\vfill\eject

\subsection{Corrections to Section 4.1.3 of Wand (2017)}

The iterated Inverse G-Wishart fragment was introduced in Section 4.1.3 of Wand (2017) 
and  it is one of the five fundamental fragments of semiparametric regression given
in Table 1. However, there are some errors due to the author of Wand (2017) failing
to recognise particular subtleties regarding the Inverse G-Wishart distribution,
as discussed in Section \ref{sec:IGWdistn}. We now point out misleading or
erroneous aspects in Section 4.1.3 of Wand (2017).

Firstly, in Wand (2017) $\bTheta_1$ plays the role of $\bSigma$ and $\bTheta_2$ plays
the role of $\bA$. The dimension of $\bTheta_1$ and $\bTheta_2$ is denoted by $d^{\Theta}$.
The first displayed equation of Section 4.1.3 is 
\begin{equation}
\bTheta_1|\bTheta_2\sim\mbox{Inverse-G-Wishart}(G,\kappa,\bTheta_2^{-1})
\label{eq:XmasEve}
\end{equation}
for $\kappa>d^{\Theta}-1$ but it is only in the $G=\Gfull$ case that such
a statement is reasonable for general $d^{\Theta}\in\naturalNumbers$. When $G=\Gfull$ 
then $\kappa=\xi-d^{\Theta}+1$ according the notation used in the current article.
Therefore, (\ref{eq:XmasEve}) involves a different parameterisation to that
used throughout this article. Therefore, our first correction is to
replace the first displayed equation of Section 4.1.3 of Wand (2017) by:
$$\bTheta_1|\bTheta_2\sim\mbox{Inverse-G-Wishart}(G,\xi,\bTheta_2^{-1})$$
where $\xi>0$ if $G=\Gdiag$ and $\xi>2d^{\Theta}-2$ if $G=\Gfull$.

The following sentence in Section 4.1.3 of Wand (2017):
``The fragment factor is of the form
$$\pDens(\bTheta_1|\bTheta_2)\propto|\bTheta_2|^{-\kappa/2}
|\bTheta_1|^{-(\kappa+d^{\Theta}+1)/2}
\exp\left\{-\smhalf\mbox{tr}(\bTheta_1^{-1}\bTheta_2^{-1})\right\}\mbox{\null''}
$$
should instead be
``The fragment factor is of the form
$$\pDens(\bTheta_1|\bTheta_2)\propto
\left\{
\begin{array}{l}
|\bTheta_2|^{-(\xi-d^{\Theta}+1)/2}
|\bTheta_1|^{-(\xi+2)/2}
\exp\left\{-\smhalf\mbox{tr}(\bTheta_1^{-1}\bTheta_2^{-1})\right\}
\quad\mbox{if $G=\Gfull$,}\\[1ex]
|\bTheta_2|^{-\xi/2}
|\bTheta_1|^{-(\xi+2)/2}
\exp\left\{-\smhalf\mbox{tr}(\bTheta_1^{-1}\bTheta_2^{-1})\right\}
\quad\mbox{if $G=\Gdiag$.''}\\[1ex]
\end{array}
\right.
$$

In equation (31) of Wand (2017), the first entry
of the vector on the right-hand side of the $\longleftarrow$ should 
be 
$$-(\xi+2)/2\quad\mbox{rather than}\quad -(\kappa+d^{\Theta}+1)/2.$$
To match the correct parameterisation of the Inverse G-Wishart distribution,
as used in the current article, equation (32) of Wand (2017)
should be
$$\mbox{\null``}E(\bX^{-1})\quad\mbox{where}\quad 
\bX\sim\mbox{Inverse-G-Wishart}(G,\xi,\bLambda)\mbox{\null''}.$$

The equation in Section 4.1.3 of Wand (2017):
$$
\mbox{\null``}
\etaSUBpThetaOneThetaTwoTOThetaTwo\thickarrow
\left[
\begin{array}{c}
-\kappa/2\\[1ex]
-\smhalf\vecof\Big(
E_{\mbox{\footnotesize$p(\bTheta_1|\bTheta_2)\to\bTheta_2$}}
(\bTheta_1^{-1})\Big)
\end{array}
\right]\mbox{\null''}
$$
should be replaced by 
$$
\mbox{\null``}
\etaSUBpThetaOneThetaTwoTOThetaTwo\thickarrow
\left[
\begin{array}{c}
-(\xi+2-2\omega_2)/2\\[1ex]
-\smhalf\vecof\Big(
E_{\mbox{\footnotesize$p(\bTheta_1|\bTheta_2)\to\bTheta_2$}}
(\bTheta_1^{-1})\Big)
\end{array}
\right]
$$
where $\omega_2$ depends on the graph of the Inverse G-Wishart distribution corresponding to 
$E_{\mbox{\footnotesize$p(\bTheta_1|\bTheta_2)\to\bTheta_2$}}$.
If the graph is $\Gfull$ then $\omega_2=(d^{\Theta}+1)/2$ and if the graph is $\Gdiag$
then $\omega_2=1$.''

Lastly the iterated Inverse G-Wishart fragment natural parameter updates
given by equations (36) and (37) of Wand (2017) are affected by the oversights
described in the preceding paragraphs. They should be replaced by the updates 
given in Algorithm \ref{alg:IterInvGWishFrag} with $\bTheta_1=\bSigma$
and $\bTheta_2=\bA$.

\section{Use of the Fragments for Covariance Matrix Prior Specification}\label{sec:priorSpec}

The underlying rationale for the Inverse G-Wishart prior and iterated Inverse G-Wishart fragments 
is their ability to facilitate the specification of a wide range of covariance matrix priors
within the variational message passing framework. In the $d=1$ special case, covariance
matrix parameters reduce to variance parameters and their square roots are standard deviation 
parameters. In this section we spell out how the fragments, and their natural parameter updates
in Algorithms \ref{alg:InvGWishPriorFrag} and \ref{alg:IterInvGWishFrag}, can be used
for prior specification in important special cases.

\subsection{Imposing an Inverse Chi-Squared Prior on a Variance Parameter}

Let $\sigma^2$ be a variance parameter and consider the prior imposition
$$\sigma^2\sim\mbox{Inverse-$\chi^2$}(\delta_{\sigma^2},\lambda_{\sigma^2})$$
for hyperparameters $\delta_{\sigma^2},\lambda_{\sigma^2}>0$,
within a variational message passing scheme. Then Algorithm 
\ref{alg:InvGWishPriorFrag} should be called with inputs set to:
$$\GsubTheta=\Gfull,\quad \xiSubTheta=\delta_{\sigma^2},\quad 
\LambdaSubTheta=\lambda_{\sigma^2}.$$

\subsection{Imposing an Inverse Gamma Prior on a Variance Parameter}

Let $\sigma^2$ be a variance parameter and consider the prior imposition
\begin{equation}
\sigma^2\sim\mbox{Inverse-Gamma}(\alphasigsq,\betasigsq)
\label{eq:IGpriorSpec}
\end{equation}
for hyperparameters $\alphasigsq,\betasigsq>0$. 
The density function corresponding to (\ref{eq:IGpriorSpec}) is
$$p(\sigma^2;\alphasigsq,\betasigsq)\propto (\sigma^2)^{-\alphasigsq-1}
\exp\{-\betasigsq/(\sigma^2)\}I(\sigma^2>0).$$
Note that the Inverse Chi-Squared and Inverse Gamma distributions
are simple reparameterisations of each other since
$$x\sim\mbox{Inverse-$\chi^2$}(\delta,\lambda)
\quad\mbox{if and only if}\quad
x\sim\mbox{Inverse-Gamma}\big(\smhalf\delta,\smhalf\lambda\big).
$$
To achieve (\ref{eq:IGpriorSpec}) Algorithm \ref{alg:InvGWishPriorFrag} should be called with inputs set to:
$$\GsubTheta=\Gfull,\quad \xiSubTheta=2\alphasigsq,\quad 
\LambdaSubTheta=2\betasigsq.$$

\subsection{Imposing an Inverse Wishart Prior on a Covariance Matrix Parameter}

A random matrix $\bX$ is defined to have an Inverse Wishart distribution
with shape parameter $\kappa$ and scale matrix $\bSigma$,
written $\bX\sim\mbox{Inverse-Wishart}(\kappa,\bLambda)$,
if and only if the density function of $\bX$ is
\begin{equation}
\begin{array}{rcl}
	\pDens(\bX)&=&\displaystyle{\frac{|\bLambda|^{\kappa/2}}{2^{\kappa d/2}\pi^{d(d-1)/4}
		\prod_{j=1}^d\Gamma(\frac{\kappa+1-j}{2})}}\,
	|\bX|^{-(\kappa+d+1)/2}\exp\{-\smhalf\mbox{\rm tr}(\bLambda\bX^{-1})\}\\[2ex]
	&&\quad\times 
	I(\bX\ \mbox{a symmetric and positive definite $d\times d$ matrix}).
\end{array}
\label{eq:GelmanTable}
\end{equation}
Note that this is the common parameterisation of the Inverse Wishart distribution
(e.g. Table A.1 of Gelman \textit{et al.}, 2014). 
Crucially, (\ref{eq:GelmanTable}) uses  a \emph{different} shape parametrization 
from that used for the Inverse G-Wishart distribution in Definition \ref{def:IGWdefn} 
when $G=\Gfull$ with the relationship between the two shape
parameters given by $\kappa=\xi-d+1$. Even though the more general Inverse G-Wishart
family is important for the internal workings of variational message passing,
the ordinary Inverse Wishart distribution, with the parameterisation 
as given in (\ref{eq:GelmanTable}), is more common when 
imposing a prior on a covariance matrix.

Let $\bSigma$ be a $d\times d$ matrix and consider the prior imposition
\begin{equation}
\bSigma\sim\mbox{Inverse-Wishart}(\kappaSubSigma,\LambdaSubSigma)
\label{eq:classicalIW}
\end{equation}
for hyperparameters $\kappaSubSigma,\LambdaSubSigma>0$,
within a variational message passing scheme. 
Then Algorithm 
\ref{alg:InvGWishPriorFrag} should be called with inputs set to:
$$\GsubTheta=\Gfull,\quad \xiSubTheta=\kappaSubSigma+d-1,\quad 
\LambdaSubTheta=\LambdaSubSigma.$$

\subsection{Imposing a Half-$t$ Prior on a Standard Deviation Parameter}

Consider the prior imposition
\begin{equation}
\sigma\sim\mbox{Half-$t$}(\ssigma,\nusigma)
\label{eq:sigmaHt}
\end{equation}
for a scale parameter $\ssigma>0$ and a degrees of freedom parameter
$\nusigma>0$.
The density function corresponding to (\ref{eq:sigmaHt})
is such that $\pDens(\sigma)\propto \{1+(\sigma/\ssigma)^2/\nusigma\}^{-(\nusigma+1)/2}I(\sigma>0)$.
This is equivalent to 
\begin{equation}
\sigma^2|a\sim\mbox{Inverse-$\chi^2$}(\nusigma,1/a)\quad\mbox{and}\quad
a\sim\mbox{Inverse-$\chi^2$}(1,1/\ssigma^2).
\label{eq:HtPriorAux}
\end{equation}
Since $d=1$, the graphs $\Gfull$ and $\Gdiag$ are the same -- 
a single node graph. Treating $\sigma^2$ and $a$ as $1\times1$ matrices 
we can re-write (\ref{eq:HtPriorAux}) as
$$\sigma^2|a\sim\mbox{Inverse-G-Wishart}(\Gfull,\nusigma,a^{-1})\quad\mbox{and}\quad
a\sim\mbox{Inverse-G-Wishart}(\Gdiag,1,(\nusigma\,\ssigma^2)^{-1})
$$
(e.g. Armagan \textit{et al.}, 2011). The specification 
$$a\sim\mbox{Inverse-G-Wishart}(\Gdiag,1,(\nusigma\ssigma^2)^{-1})$$
involves calling Algorithm \ref{alg:InvGWishPriorFrag}
with 
$$\GsubTheta=\Gdiag,\quad\xiSubTheta=1\quad\mbox{and}\quad
\LambdaSubTheta=(\nusigma\,\ssigma^2)^{-1}.$$
The output is the single node graph $\GSUBpThetaTOTheta$
and the $2\times1$ natural parameter vector 
$$\etaSUBpThetaTOTheta=\biggerbdeta_{\pDens(a)\rightarrow a}.$$

The specification 
$$\sigma^2|a\sim\mbox{Inverse-G-Wishart}(\Gfull,\nusigma,a^{-1})$$
implies that Algorithm \ref{alg:IterInvGWishFrag} is called 
with graph input $G=\Gfull$, shape parameter input 
$\xi=\nusigma$ and message parameter inputs
$$\etaSUBpSigmaATOSigma=\biggerbdeta_{\pDens(\sigma^2|a)\to\sigma^2},
\quad\etaSUBSigmaTOpSigmaA=\biggerbdeta_{\sigma^2\to \pDens(\sigma^2|a)},$$
and
$$\GSUBpSigmaATOA=\Gdiag,\quad\etaSUBpSigmaATOA=\biggerbdeta_{\pDens(\sigma^2|a)\to a}
\quad\mbox{and}\quad\etaSUBATOpSigmaA=\biggerbdeta_{a\to \pDens(\sigma^2|a)}.$$
Note that in this $d=1$ special case $\Gfull$ and $\Gdiag$ are both the single node graph.

\subsubsection{The Half-Cauchy Special Case}

The special case of 
\begin{equation}
\sigma\sim\mbox{Half-Cauchy}(\ssigma).
\label{eq:sigmaHC}
\end{equation}
corresponds to $\nusigma=1$. 
The density function corresponding to (\ref{eq:sigmaHC})
is such that $\pDens(\sigma)\propto \{1+(\sigma/\ssigma)^2\}^{-1}I(\sigma>0)$.
Therefore, one should set $\xi=1$ in the call to Algorithm \ref{alg:IterInvGWishFrag}.

\subsection{Imposing a Huang-Wand Prior on a Covariance Matrix}

To impose the Huang-Wand prior 
$$\bSigma\sim\mbox{Huang-Wand}(\sSigmaOne,\ldots,\sSigmad)$$
in a variational message passing framework we should have the inputs to 
Algorithm \ref{alg:InvGWishPriorFrag} being as follows:
$$\GsubTheta=\Gdiag,\quad\xiSubTheta=1
\quad\mbox{and}\quad\LambdaSubTheta=\big\{2\,\diag(\sSigmaOne^2,\ldots,\sSigmad^2)\big\}^{-1}.$$
The graph parameter input to Algorithm \ref{alg:IterInvGWishFrag} should be 
$G=\Gfull$ and the shape parameter input should be $\xi=2d$.

\subsection{Imposing a Matrix-$F$ Prior on a Covariance Matrix}

To impose the Matrix-$F$ prior
$$\bSigma\sim F(\nuMP,\deltaMP,\bBMP)$$
in a variational message passing framework the inputs to 
Algorithm \ref{alg:InvGWishPriorFrag} should be as follows:
$$\GsubTheta=\Gfull,\quad\xiSubTheta=\nuMP+d-1
\quad\mbox{and}\quad\LambdaSubTheta=\bBMP^{-1}.$$
The graph parameter input to Algorithm \ref{alg:IterInvGWishFrag} should be 
$G=\Gfull$ and the shape parameter input should be $\xi=\deltaMP+2d-2$.

\subsection{Tabular Summary of Fragment-Based Prior Specification}

Table \ref{tab:fragPrior} summarizes the results of this section and
is a crucial reference for placing priors of covariance matrix,
variance and standard deviation parameters in variational message passing
schemes that make use of Algorithms \ref{alg:InvGWishPriorFrag} and 
\ref{alg:IterInvGWishFrag}.

\begin{table}[ht]
\begin{center}
{\setlength{\tabcolsep}{4pt}
\begin{tabular}{lccccccc}
\hline
\multicolumn{1}{c}{\null}&
\multicolumn{3}{c}{Algorithm 1}& 
\multicolumn{1}{c}{\null}&
\multicolumn{3}{c}{Algorithm 2}\\
\cmidrule(r){2-4}\cmidrule(l){6-8}
\multicolumn{1}{c}{prior specification}& 
\multicolumn{1}{c}{$\GsubTheta$} & 
\multicolumn{1}{c}{$\xiSubTheta$}& 
\multicolumn{1}{c}{$\LambdaSubTheta$}&
\multicolumn{1}{c}{\null}& 
\multicolumn{1}{c}{$\xi$}& 
\multicolumn{1}{c}{$G$}& 
\multicolumn{1}{c}{$\GSUBATOpSigmaATAB$ }
\\
\hline\\[-1.5ex]
\begin{small}$\sigma^2\sim\mbox{Inverse-$\chi^2$}(\delta_{\sigma^2},\lambda_{\sigma^2})$\end{small} 
& $\Gfull$ & $\delta_{\sigma^2}$   & $\lambda_{\sigma^2}$&\null&   N.A.& N.A. & N.A. \\[1ex]
\begin{small}$\sigma^2\sim\mbox{Inv.-Gamma}(\alphasigsq,\betasigsq)$\end{small}           
& $\Gfull$ & $2\alphasigsq$& $2\betasigsq$ &\null &  N.A.& N.A. & N.A. \\[1ex]
\begin{small}$\bSigma\sim\mbox{Inv.-Wishart}(\kappaSubSigma,\LambdaSubSigma)$\end{small} 
& $\Gfull$ & $\kappaSubSigma+d-1$   & $\LambdaSubSigma$ &\null &   N.A.& N.A. & N.A. \\[1ex]
\begin{small}$\sigma\sim\mbox{Half-$t$}(\ssigma,\nusigma)$\end{small}                               
& $\Gdiag$ & $1$   & $(\nusigma\ssigma^2)^{-1}$ &\null & $\nusigma$    & $\Gfull$  &  $\Gdiag$\\[1ex]
\begin{small}$\sigma\sim\mbox{Half-Cauchy}(\ssigma)$\end{small}                                 
& $\Gdiag$ & $1$   & $(\ssigma^2)^{-1}$ &\null &  $1$    & $\Gfull$  &  $\Gdiag$\\[1ex]
\begin{small}$\bSigma\sim\mbox{Huang-Wand}$ \end{small}           
& $\Gdiag$ & $1$   & $\big\{2\,\diag(\sSigmaOne^2,$ &\null &  $2d$    & $\Gfull$  &  $\Gdiag$\\[0ex]  
$\qquad\ (\sSigmaOne,\ldots,\sSigmad)$               &   &       &      $\null\ \ \ldots,\sSigmad^2)\big\}^{-1}$ 
&\null &       &       &     \\[1ex]  
\begin{small}$\bSigma\sim\mbox{Matrix-$F$}$ \end{small}           
&$\Gfull$ & $\nuMP+$  & $\bBMP^{-1}$ &\null &  $\deltaMP+$ & $\Gfull$  &  $\Gfull$\\[0ex]  
$\qquad\ (\nuMP,\deltaMP,\bBMP)$               &   & $\ d-1$      &   
&\null & $\ 2d-2$       &       &     \\[1ex]  
\hline
\end{tabular}
}
\end{center}
\caption{\it Specifications of inputs of Algorithms \ref{alg:InvGWishPriorFrag} 
and \ref{alg:IterInvGWishFrag} for several variance, standard deviation and 
covariance matrix prior impositions. The abbreviation N.A. stands for not 
applicable since Algorithm \ref{alg:IterInvGWishFrag} is not needed for the 
first three prior impositions.}
\label{tab:fragPrior} 
\end{table}

\section{Illustrative Example}\label{sec:illustrative}

We illustrate the use of Algorithms \ref{alg:InvGWishPriorFrag} 
and \ref{alg:IterInvGWishFrag} for the case of Bayesian linear 
mixed models with $t$ distribution responses. Such $t$-based models
impose a form of robustness in situations where the responses are susceptible
to having outlying values (e.g. Lange \textit{et al.}, 1989).
The notation $y\sim t(\mu,\sigma,\nu)$ indicates that the random
variable $y$ has a $t$ distribution with location parameter $\mu$,
scale parameter $\sigma>0$ and degrees of freedom parameter $\nu>0$.
The corresponding density function of $y$ is 
$$\pDens(y)=\displaystyle{\frac{\Gamma\left(\frac{\nu+1}{2}\right)}
{\sigma\sqrt{\pi\nu}\Gamma(\nu/2)[1+\{(y-\mu)/\sigma\}^2/\nu]^{\frac{\nu+1}{2}}}}.
$$
Now suppose that the response data consists of repeated measures 
within each of $m$ groups. Let 
$$y_{ij}\equiv\mbox{the $j$th response for the $i$th group},\quad 1\le j\le n_i,\ 1\le i\le m,$$
and then let $\by_i$, $1\le i\le m$, be the $n_i\times 1$ vectors containing $y_{ij}$ data
for the $i$th group. For each $1\le i\le m$, let $\bX_i$ be $n_i\times p$ design matrices 
corresponding to the fixed effects and $\bZ_i$ be $n_i\times q$ design matrices corresponding 
to the random effects. Next put
\begin{equation}
\by\equiv\left[  
\begin{array}{c}
\by_1\\
\vdots\\
\by_m
\end{array}
\right],
\quad 
\bX\equiv\left[  
\begin{array}{c}
\bX_1\\
\vdots\\
\bX_m
\end{array}
\right]
\quad\mbox{and}\quad 
\bZ\equiv\blockdiag{1\le i\le m}(\bZ_i)
\label{eq:yXZdefn}
\end{equation}
and define $N=n_1+\ldots+n_m$ to be the number of rows in each of $\by$, $\bX$ and $\bZ$.
Let $y_{\ell}$ be the $\ell$th entry of $\by$, $1\le\ell\le N$.
The family of Bayesian $t$ response linear mixed models that we consider is
\begin{equation}
\begin{array}{c}
y_{\ell}|\bbeta,\bu,\sigma\simind t\big((\bX\bbeta+\bZ\bu)_{\ell},\sigma,\nu\big),
\quad 1\le \ell\le N,\quad \bu|\bSigma\sim N(\bzero,\bI_m\otimes\bSigma),\\[2ex]
\bbeta\sim N(0,\sigma_{\bbeta}^2\bI),\quad
\sigma\sim\mbox{Half-Cauchy}(\ssigma),\quad \smhalf\nu\sim\mbox{Moon-Rock}(0,\lambda_{\nu}),\\[2ex]
\bSigma\sim\mbox{Huang-Wand}\big(\sSigmaOne,\ldots,\sSigmaq\big)
\end{array}
\label{eq:tRespModel}
\end{equation}
for hyperparameters 
$\sigma_{\bbeta},\ssigma,\lambda_{\nu},\sSigmaOne,\ldots,\sSigmaq>0$.

As explained in McLean \myand Wand (2019), the Moon Rock family of distributions is 
conjugate for the parameter $\smhalf\nu$, with the notation $x\sim\mbox{Moon-Rock}(\alpha,\beta)$
indicating that the corresponding density function satisfies
$\pDens(x)\propto\{x^x/\Gamma(x)\}^{\alpha}\exp(-\beta x)I(x>0)$.
In the variational message passing treatment of the degrees of freedom parameter
it is simpler to work with 
$$\halfnu\equiv \smhalf\nu\quad\mbox{so that}\quad\halfnu\sim\mbox{Moon-Rock}(0,\lambda_{\nu}).$$
After the approximate posterior density function of $\halfnu$ is
obtained via variational message passing, it is trivial to then 
obtain the same for $\nu$. Hence, we work with $\halfnu$, 
rather than $\nu$, in the upcoming description of variational
message passing-based fitting and inference for (\ref{eq:tRespModel}).

\begin{figure}[h]
\centering
{\includegraphics[width=0.8\textwidth]{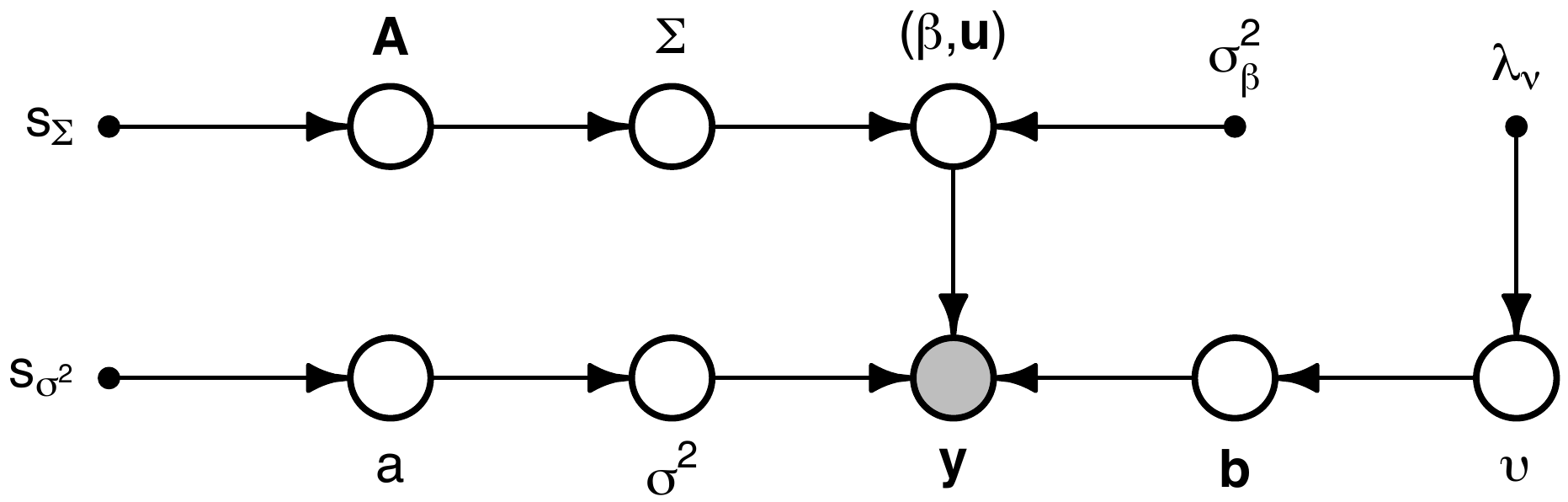}}
\caption{\it Directed acyclic graph corresponding to the
$t$ response linear mixed model (\ref{eq:tRespModel}) with auxiliary
variable representations (\ref{eq:auxVarOne})--(\ref{eq:auxVarThree}).
The shaded circle corresponds to the observed data. The unshaded circles
correspond to model parameters and auxiliary variables. The small
solid circles correspond to hyperparameters.
}
\label{fig:tMixModDAG} 
\end{figure}

Next note that
$$\by_{\ell}|\bbeta,\bu,\sigma\simind t\Big((\bX\bbeta+\bZ\bu)_{\ell},\sigma,2\halfnu\Big),
\quad 1\le \ell\le N$$
is equivalent to 
\begin{equation}
\by_{\ell}|\bbeta,\bu,\sigma^2,b_{\ell}\sim 
N\big((\bX\bbeta+\bZ\bu)_{\ell},b_{\ell}\sigma^2\big),
\quad b_{\ell}|\halfnu\simind\mbox{Inverse-$\chi^2$}\left(2\halfnu,2\halfnu\right),
\label{eq:auxVarOne}
\end{equation}
$\sigma\sim\mbox{Half-Cauchy}(\ssigma)$ is equivalent to 
\begin{equation}
\sigma^2|a\sim\mbox{Inverse-$\chi^2$}(1,1/a)\quad\mbox{and}\quad
a\sim\mbox{Inverse-$\chi^2$}(1,1/\ssigma^2)
\label{eq:auxVarTwo}
\end{equation}
and $\bSigma\sim\mbox{Huang-Wand}\big(\sSigmaOne,\ldots,\sSigmaq\big)$
is equivalent to 
\begin{equation}
\begin{array}{c}
\bSigma|\bA\sim\mbox{Inverse-G-Wishart}(\Gfull,2q,\bA^{-1}),\\[1ex]
\bA\sim\mbox{Inverse-G-Wishart}
\Big(\Gdiag,1,\big\{2\,\diag(\sSigmaOne^2,\ldots,\sSigmaq^2)\big\}^{-1}\Big).
\end{array}
\label{eq:auxVarThree}
\end{equation}
Substitution of (\ref{eq:auxVarOne}), (\ref{eq:auxVarTwo}) and (\ref{eq:auxVarThree})
into (\ref{eq:tRespModel}) leads to the hierarchical Bayesian model depicted
as a directed acyclic graph in Figure \ref{fig:tMixModDAG} with $\bb\equiv(b_1,\ldots,b_N)$.
The unshaded circles in Figure \ref{fig:tMixModDAG} correspond to model parameters
and auxiliary variables and will be referred to as \emph{hidden nodes}.

\begin{figure}[h]
\centering
{\includegraphics[width=1\textwidth]{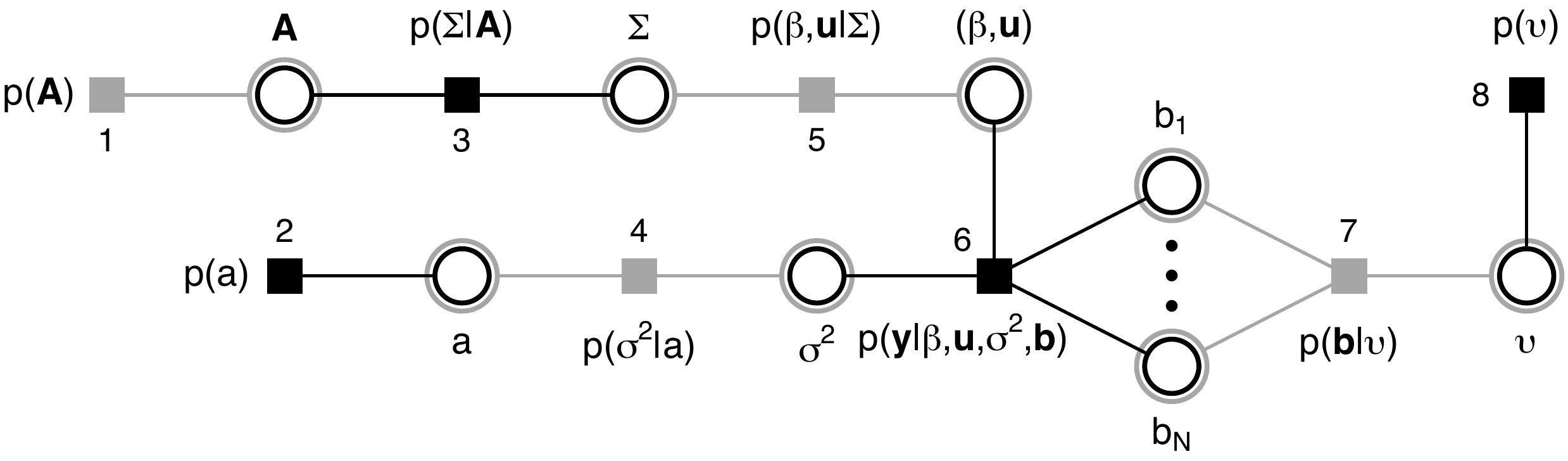}}
\caption{\it Factor graph corresponding to the
$t$ response linear mixed model (\ref{eq:tRespModel}) with auxiliary
variable representations (\ref{eq:auxVarOne})--(\ref{eq:auxVarThree}).
The circular nodes correspond to stochastic nodes in the 
$\qDens$-density factorization in (\ref{eq:finalMF}). The rectangular
nodes correspond to the factors on the right-hand side of (\ref{eq:fullFactor}).
The fragments are numbered  $1$ to $8$ according to appearance from left to right. 
Shading is used to show the distinction between adjacent fragments.
}
\label{fig:tMixModFacGraph} 
\end{figure}

Consider the following mean field approximation of the joint 
posterior of the hidden nodes in Figure \ref{fig:tMixModDAG}
\begin{equation}
\pDens(\bbeta,\bu,\sigma^2,\halfnu,\bSigma,a,\bA,\bb|\by)
\approx \qDens(\bbeta,\bu,a,\bA,\bb)\qDens(\sigma^2,\bSigma,\halfnu)
\label{eq:lessStringent}
\end{equation}
where $\qDens$ denotes the approximate posterior density functions 
of the relevant parameters. Application of induced factor results 
(e.g. Bishop, 2006; Section 10.2.5) leads to the additional factorizations
$$\qDens(\bbeta,\bu,a,\bA,\bb)=\qDens(\bbeta,\bu)\qDens(a)\qDens(\bA)\prod_{\ell=1}^N\qDens(b_{\ell})
\quad\mbox{and}\quad
\qDens(\sigma^2,\bSigma,\halfnu)=\qDens(\sigma^2)\qDens(\bSigma)\qDens(\halfnu).
$$
and so the restriction given in (\ref{eq:lessStringent}) is equivalent to 
\begin{equation}
\pDens(\bbeta,\bu,\sigma^2,\halfnu,\bSigma,a,\bA,\bb|\by)
\approx \qDens(\bbeta,\bu)\qDens(\sigma^2)\qDens(\halfnu)\qDens(\bSigma)\qDens(a)\qDens(\bA)
\prod_{\ell=1}^N\qDens(b_{\ell}).
\label{eq:finalMF}
\end{equation}
Figure \ref{fig:tMixModFacGraph} is a factor graph representation of 
the joint density function of all random variables and vectors,
or stochastic nodes, in
Figure \ref{fig:tMixModDAG} hierarchical model, with unshaded circles 
for each stochastic node according to the $\qDens$-density factorization
given in (\ref{eq:finalMF}) and filled-in rectangles corresponding
to factors on the right-hand side of
\begin{equation}
\begin{array}{l}
\pDens(\by,\bbeta,\bu,\sigma^2,\halfnu,\bSigma,a,\bA,\bb)\\[1ex]
\qquad\qquad
=\pDens(\bA)\pDens(a)\pDens(\bSigma|\bA)\pDens(\sigma^2|a)
\pDens(\bbeta,\bu|\bSigma)\pDens(\by|\bbeta,\bu,\sigma^2,\bb)
\pDens(\bb|\halfnu)\pDens(\halfnu).
\end{array}
\label{eq:fullFactor}
\end{equation}
Edges join each factor to a stochastic node that appears in the factor.
To aid upcoming discussion, the fragments are numbered $1$ to $8$
according to appearance from left to right. Recall that a fragment 
is a sub-graph consisting of a factor and all of its neighboring nodes. 
Figure \ref{fig:tMixModFacGraph} uses shading to show the distinction
between adjacent fragments.

Note that (e.g. Minka, 2005; Wand, 2017) the variational message passing iteration 
loop has the following generic steps:
$$
\begin{array}{lll}
&\mbox{1. Choose a factor.}\\
&\mbox{2. Update the parameter vectors of the messages passed from}\\
&\mbox{\ \ \ \ the factor's neighboring stochastic nodes to the factor.}\\
&\mbox{3. Update the parameter vectors of the messages passed}\\
&\mbox{\ \ \ \ from the factor to its neighboring stochastic nodes.}
\end{array}
$$
Step 2. is very simple and has generic form given by, for example, (7) of Wand (2017).
In the Figure \ref{fig:tMixModFacGraph} factor graph an example of Step 2. is:
\begin{equation}
{\setlength\arraycolsep{3pt}
\begin{array}{rcl}
&&\mbox{the message passed from $\bSigma$ to $p(\bbeta,\bu|\bSigma)$}\\[1ex]
&&\qquad\qquad=\mbox{the message passed from $p(\bSigma|\bA)$ to $\bSigma$ 
in the previous iteration.}
\end{array}
}
\label{eq:StoFexamp}
\end{equation}
In terms of natural parameter vector updates, (\ref{eq:StoFexamp}) corresponds to:
$$\etaSUBSigmaTOpSigmaA\thickarrow\etaSUBpSigmaATOSigma.$$
Most of the other stochastic node to factor updates
in Figure \ref{fig:tMixModFacGraph} have an analogous form.
The exception are the messages passed within fragments 6 and 7,
which require use of the slightly more complicated form as
given by, for example, equation (7) of Wand (2017).

It remains to discuss Step 3., corresponding to the factor to stochastic
node updates:
\begin{itemize}
\item Fragments 1 and 2 are Inverse G-Wishart prior fragments and the
factor to stochastic node parameter vector updates are performed according
to Algorithm \ref{alg:InvGWishPriorFrag}. In view of Table \ref{tab:fragPrior}, 
the graph and shape hyperparameter inputs are 
$\GsubTheta=\Gdiag$ and $\xiSubTheta=1$. For fragment 1 the 
rate hyperparameter is $\LambdaSubTheta
=\{2\diag\big(\sSigmaOne^2,\ldots,\sSigmaq^2\big)\}^{-1}$.
For fragment 2 the rate hyperparameter
is $\LambdaSubTheta=(s^2_{\sigma})^{-1}$. 
\item Fragments 3 and 4 are iterated Inverse G-Wishart prior fragments and the
factor to stochastic node parameter vector updates are performed according
to Algorithm \ref{alg:IterInvGWishFrag}. As shown in Table \ref{tab:fragPrior},
the graph inputs should be
$$\GSUBATOpSigmaA=\Gdiag,\ \GSUBaTOpsigsqa=\Gdiag,\ \GSUBSigmaTOpSigmaA=\Gfull,
\ \mbox{and}\ \GSUBsigsqTOpsigsqa=\Gfull.
$$
The first two of these are imposed by the messages passed from fragments 1 and 2.
For fragment 3, the shape parameter input is $\xi=2q$. 
For fragment 4, the shape parameter input is $\xi=1$. 
\item Fragment 5 is the Gaussian penalization fragment described
in Section 4.1.4 of Wand (2017) with, in the notation given there, $L=1$,
$\bmu_{\btheta_0}=\bzero$ and $\bSigma_{\btheta_0}=\sigma_{\bbeta}^2\bI$.
\item Fragments 6 and 7 correspond to the $t$ likelihood fragment.
Its natural parameter updates are provided by Algorithm 2 of
McLean \myand Wand (2019). 
\item Fragment 8 corresponds to the imposition of a Moon Rock
prior distribution on a shape parameter. This is a very simple
fragment for which the only inputs are the Moon Rock prior 
specification hyperparameters and the output is the 
natural parameter vector of the Moon Rock prior density function.
Since this fragment is not listed as an algorithm in this article
or elsewhere, we provide further details in the paragraph
after the next one.
\end{itemize}

For Fragments 5, 6 and 7 simple conversions between two different
versions of natural parameter vectors need to be made.
Section \ref{sec:vecANDvech} of the web-supplement
explains these conversions.

The most general Moon Rock prior specification for a generic parameter $\theta$ is 
$$\theta\sim\mbox{Moon-Rock}(\alpha_{\theta},\beta_{\theta}).$$
This corresponds to the prior density function having exponential family form
$$p(\theta)\propto\exp\left\{
\left[
\begin{array}{c}
\theta\log(\theta)-\log\Gamma(\theta)\\[1ex]
\theta
\end{array}
\right]^T
\left[                
\begin{array}{c}
\alpha_{\theta}\\[1ex]
-\beta_{\theta}
\end{array}
\right]
\right\}
$$
The inputs of the Moon Rock prior fragment are $\alpha_{\theta}\ge 0$ and $\beta_{\theta}>0$
and the output is the natural parameter vector 
$$\etaSUBpthetaTOtheta\thickarrow\left[                
\begin{array}{c}
\alpha_{\theta}\\[1ex]
-\beta_{\theta}
\end{array}
\right].
$$
Since, for the $t$ response mixed model illustrative example, 
we have the prior imposition 
$\halfnu\sim\mbox{Moon-Rock}(0,\lambda_{\nu})$
we simply call the Moon Rock prior fragment with 
$(\alpha_{\theta},\beta_{\theta})$ set to $(0,\lambda_{\nu})$.

To demonstrate variational message passing for fitting
and inference for model (\ref{eq:tRespModel}), we simulated 
data according to the dimension values $p=q=2$ and the
true parameter values
\begin{equation}
\bbetaTrue=
\left[
\begin{array}{c}
-0.58 \\[1ex]
1.89
\end{array}
\right],
\quad
\sigsqTrue=0.2,
\quad
\bSigmaTrue
=
\left[
\begin{array}{cc}
2.58 & 0.22 \\[1ex]
0.22 & 1.73
\end{array}
\right]
\quad\mbox{and}\quad
\nuTrue=1.5.
\label{eq:trueValues}
\end{equation}
The sample sizes were $m=20$, with $n_i=15$ observations per group,
and the predictor data were generated from the Uniform distribution
on the unit interval. The hyperparameter values were set at
$$\sigma_{\bbeta}=s_{\sigma}=\sSigmaOne=\sSigmaTwo=10^5\quad\mbox{and}\quad
\lambda_{\nu}=0.01.$$
We ran the variational message passing algorithm as described above
until the relative change the variational parameters was below $10^{-10}$.
as well as Markov chain Monte Carlo via
the \textsf{R} language (\textsf{R} Core Team, 2020) package 
\texttt{rstan} (Stan Development Team, 2019). 
For Markov chain Monte Carlo fitting, a warmup of size 1000
was used, followed by chains of size 5000 
retained for inference.

\begin{figure}[h]
\centering
{\includegraphics[width=0.95\textwidth]{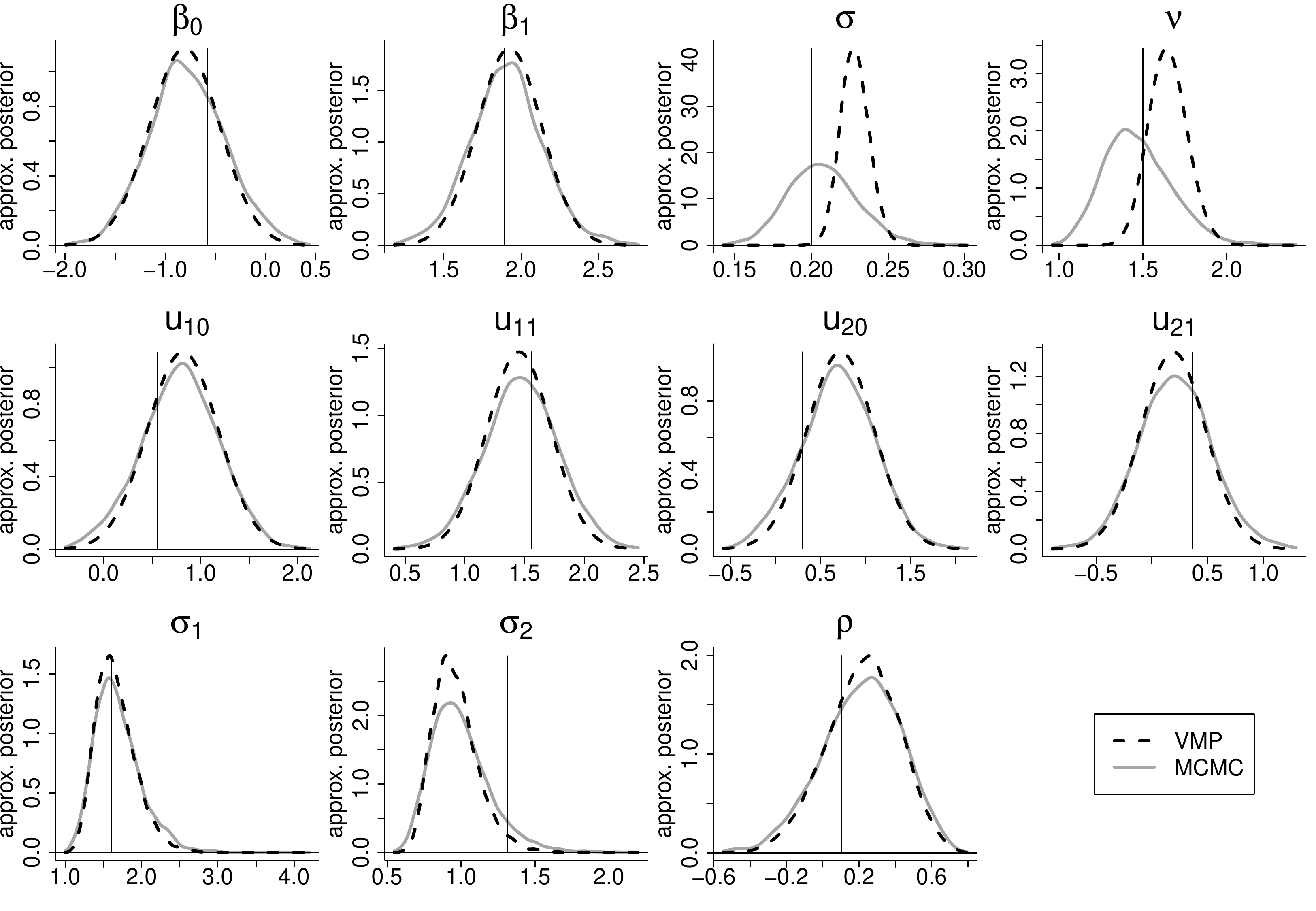}}
\caption{\it Approximate posterior density functions for the parameters
in model (\ref{eq:tRespModel}) based on both variational message
passing (VMP) and Markov chain Monte Carlo (MCMC) algorithms applied to 
data simulated according to the true values (\ref{eq:trueValues})
and sample sizes, predictor values and hyperparameter values
as described in the text. The vertical lines indicate true parameter values. 
}
\label{fig:tMixModPosterDens} 
\end{figure}

Figure \ref{fig:tMixModPosterDens} compares the approximate
posterior density functions based on both variational
message passing (VMP) and Markov chain Monte Carlo (MCMC).
The middle row performs the comparison for the 
random intercept and slope parameters, $u_{i0}$ and $u_{i1}$,
for $i=1,2$. The parameters in the third row of Figure \ref{fig:tMixModPosterDens} 
are for the standard deviation and correlation parameters in the $\bSigma$ 
matrix, according to the notation $(\bSigma)_{11}=\sigma_1$, $(\bSigma)_{22}=\sigma_2$
and $(\bSigma)_{12}=\sigma_1\sigma_2\rho$. For most of the stochastic nodes,
the accuracy of variational message passing is seen to be very good.
For $\sigma$ and $\nu$, some under-approximation of the spread and locational shift 
is apparent. A likely root cause is the imposition of the product restriction
$\qDens(\sigma,\nu)=\qDens(\sigma)\qDens(\nu)$ even though these two parameters
have a significant amount of posterior dependence.

We have prepared a bundle of \textsf{R} language
code that carries out variational message passing for this illustrative example,
including use of Algorithms \ref{alg:InvGWishPriorFrag}
and \ref{alg:IterInvGWishFrag} for the imposition
of Half Cauchy and Huang-Wand priors. This code is part of 
the web-supplement for this article.

Lastly, we point out that this illustrative example does not involve
matrix algebraic streamlining for random effects models. This relatively
new area for variational message passing research, which streamlines
calculations involving sparse matrix forms that arise in linear mixed models,
is described in Nolan, Menictas \myand Wand (2020).

\section{Closing Remarks}\label{sec:closing}

Algorithms \ref{alg:InvGWishPriorFrag} and, especially, Algorithm \ref{alg:IterInvGWishFrag} 
and their underpinnings are quite involved and dependent upon a careful study of particular 
special cases of the inverses of G-Wishart random matrices. The amount of detail
provided by this article is tedious, but necessary, to ensure that the 
fragment updates based on a single distributional
structure, the Inverse G-Wishart distribution with $G\in\{\Gfull,\Gdiag\}$,
are correct. The good news is that these algorithms only need to be derived once. 
Their implementations, within a suite of computer 
programmes for carrying out variational message passing for models containing 
variance and covariance matrix parameters, can be isolated into subroutines 
which, once working as intended, do not have
to be revisited ever again. Given the quintessence of variance and covariance
parameters in throughout statistics and machine learning, 
Algorithms \ref{alg:InvGWishPriorFrag} and Algorithm \ref{alg:IterInvGWishFrag} 
are important and fundamental contributions to variational message passing.

\section*{Acknowledgements}

We are grateful to two referees for their comments and suggestions.
This research was supported by Australian Research Council Discovery
Project DP140100441.

\section*{References}

\bib
Armagan, A., Dunson, D.B. and Clyde, M. (2011). Generalized beta mixtures of Gaussians. 
In \textit{Advances in Neural Information Processing Systems 24}, 
J.Shawe-Taylor, R.S. Zamel, P. Bartlett, F. Pereira and K.Q. Weinberger (eds.),
pp. 523--531.

\bib
Assaf, A.G., Li, G., Song, H. \myand Tsionas, M.G. (2019).
Modeling and forecasting regional tourism demand using the Bayesian global vector
autoregressive (BGVAR) model. 
\textit{Journal of Travel Research}, {\bf 58}, 383--397.

\bib
Attias, H. (1999). Inferring parameters and structure of latent variable
models by variational Bayes. In Laskey, K.B. and Prade, H. (editors)
\textit{Proceedings of the Fifteenth Conference
on Uncertainty in Artificial Intelligence}, pp. 21--30.
San Francisco: Morgan Kauffmann.

\bib
Atay-Kayis, A. \myand Massam, H. (2005).
A Monte Carlo method for computing marginal likelihood
in nondecomposable Gaussian graphical models.
\textit{Biometrika}, {\bf 92}, 317--335.

\bib
Bishop, C.M. (2006). {\it Pattern Recognition and Machine Learning.}
New York: Springer.

\bib
Chen, W.Y and Wand, M.P. (2020).
Factor graph fragmentization of expectation propagation. 
\textit{Journal of the Korean Statistical Society}, in press.

\bib
Conti, G., Fr\"{u}hwirth-Schnatter, S., Heckman, J.J. \myand Piatek, R. (2014).
Bayesian exploratory factor analysis. 
\textit{Journal of Econometrics}, {\bf 183}, 31--57.

\bib
Dawid, A.P. \myand Lauritzen, S.L. (1993). Hyper Markov laws in the statistical analysis of
decomposable graphical models. \textit{The Annals of Statistics}, \textbf{21}, 1272--1317.

\bib
Gelman, A. (2006). Prior distributions for variance parameters in hierarchical models. 
\textit{Bayesian Analysis}, {\bf 1}, 515--533.

\bib
Gelman, A., Carlin, J.B., Stern, H.S., Dunson, D.B., Vehtari, A. \myand Rubin, D.B. (2014). 
\textit{Bayesian Data Analysis, Third Edition}, Boca Raton, Florida: CRC Press.

\bib
Gentle, J.E. (2007). \textit{Matrix Algebra}.
New York: Springer.

\bib
Harezlak, J., Ruppert, D. \myand Wand, M.P. (2018).
{\it Semiparametric Regression with R}.
New York: Springer.  

\bib
Huang, A. \myand Wand, M.P. (2013).
Simple marginally noninformative prior distributions 
for covariance matrices. \textit{Bayesian Analysis}, 
{\bf 8}, 439--452.

\bib
Lange, K.L., Little, R.J.A. and Taylor, J.M.G. (1989).
Robust statistical modeling using the $t$-distribution.
{\it Journal of the American Statistical Association},
{\bf 84}, 881-896.

\bib
Letac, G. \myand Massam, H. (2007).
Wishart distributions for decomposable graphs.
\textit{The Annals of Statistics}, {\bf 35}, 1278--1323.

\bib
Maestrini, L., \myand Wand, M.P. (2018). Variational message passing for skew t regression. 
\emph{Stat}, {\bf7}, e196.

\bib
Magnus, J.R. \myand Neudecker, H. (1999). \emph{Matrix Differential Calculus with Applications in
Statistics and Econometrics, Revised Edition}. Chichester U.K.: Wiley

\bib
McCulloch, C.E., Searle, S.R. \myand Neuhaus, J.M. (2008). \textit{Generalized, Linear,
and Mixed Models, Second Edition}. New York: John Wiley \& Sons.

\bib
McLean, M.W. \myand Wand, M.P. (2019).
Variational message passing for elaborate response regression models.
\textit{Bayesian Analysis}, {\bf 14}, 371--398.

\bib
Minka, T.P. (2001).
Expectation propagation for approximate Bayesian inference.
In J.S. Breese \myand D. Koller (eds),
\textit{Proceedings of the Seventeenth Conference on Uncertainty in
Artificial Intelligence}, pp. 362--369.
Burlington, Massachusetts: Morgan Kaufmann.

\bib
Minka, T. (2005). 
Divergence measures and message passing.
\textit{Microsoft Research Technical Report Series}, 
{\bf MSR-TR-2005-173}, 1--17.

\bib
Muirhead, R.J. (1982). \textit{Aspects of Multivariate Statistical Theory.}
New York: John Wiley \& Sons.

\bib
Mulder, J. \myand Pericchi, L.R. (2018). The Matrix-$F$ prior for estimating and
testing covariance matrices. \textit{Bayesian Analysis}, {\bf 13}, 1193--1214.

\bib
Nolan, T.H., Menictas, M. and Wand, M.P. (2020).
Streamlined computing for variational inference with higher level 
random effects. Unpublished manuscript
available at \textit{https://arxiv.org/abs/1903.06616}.

\bib
Nolan, T.H. and Wand, M.P. (2017).
Accurate logistic variational message passing: 
algebraic and numerical details. \textit{Stat}, {\bf 6},
102--112.

\bib
Polson, N. G. \myand Scott, J. G. (2012). On the half-Cauchy prior for a global scale parameter. 
\textit{Bayesian Analysis}, {\bf 7}, 887--902.

\bib
\textsf{R} Core Team (2020). \textsf{R}: A language and environment
for statistical computing. \textsf{R} Foundation for Statistical Computing.
Vienna, Austria. \texttt{https://www.R-project.org/}

\bib
Roverato, A. (2000). Cholesky decomposition of a hyper inverse Wishart matrix.
\textit{Biometrika}, {\bf 87}, 99-112.

\bib
Stan Development Team (2019). \textsf{RStan}: the
\textsf{R} interface to \textsf{Stan}. 
\textsf{R} package version 2.19.2.
\texttt{http://mc-stan.org/.}

\bib
Uhler, C., Lenkoski, A. and Richards, D. (2018).
Exact formulas for the normalizing constants of Wishart
distributions for graphical models. \textit{The Annals of Statistics}, {\bf 46}, 90--118.

\bib
Wand, M.P. (2017).
Fast approximate inference for arbitrarily large semiparametric
regression models via message passing (with discussion).
\textit{Journal of the American Statistical Association}, 
{\bf 112}, 137--168.

\bib
Winn, J. \myand Bishop, C.M. (2005).
Variational message passing. \textit{Journal of Machine Learning Research},
{\bf 6}, 661--694.

\vfill\eject
%
%
\renewcommand{\theequation}{S.\arabic{equation}}
\renewcommand{\thesection}{S.\arabic{section}}
\renewcommand{\thetable}{S.\arabic{table}}
\setcounter{equation}{0}
\setcounter{table}{0}
\setcounter{section}{0}
\setcounter{page}{1}
\setcounter{footnote}{0}

\begin{center}

{\Large Web-supplement for:}
\vskip3mm

\centerline{\Large\bf The Inverse G-Wishart Distribution and Variational Message Passing}
\vskip7mm
\ifthenelse{\boolean{UnBlinded}}{\centerline{\normalsize\sc By L. Maestrini and M.P. Wand}
\vskip5mm
\centerline{\textit{University of Technology Sydney}}
\vskip6mm}{\null}
\end{center}

\section{Natural Parameter Versions and Mappings}\label{sec:vecANDvech}

Throughout this article we use the ``$\vech$'' versions of the 
natural parameter forms of the Multivariate Normal 
and Inverse G-Wishart distributions. However, Wand (2017) and
McLean \myand Wand (2019) used ``$\vecof$'' versions of these
distributions. The ``$\vech$'' version has the attraction
of being more compact since entries of symmetric matrices
are not duplicated. However, adoption of the ``$\vech$'' version
entails use of duplication matrices. For implementation in the \textsf{R}
language (\textsf{R} Core Team, 2020) we note that the 
function \texttt{duplication.matrix()} in the package 
\textsf{matrixcalc} (Novomestky, 2012) returns the duplication matrix
of a given order.

First we explain the two versions for the Multivariate Normal distribution.
Suppose that the $d\times1$ random vector $\bv$ has a $N(\bu,\bSigma)$
distribution. Then the density function of $\bv$ is
$$\pDens(\bv)\propto
\exp\left\{\left[
\begin{array}{c}
\bv\\
\vecof(\bv\bv^T)
\end{array}
\right]^T\bdetavvec
\right\}
=\exp\left\{\left[
\begin{array}{c}
\bv\\
\vech(\bv\bv^T)
\end{array}
\right]^T\bdetavvech
\right\}
$$
where 
$$\bdetavvec\equiv
\left[     
\begin{array}{c}
\bSigma^{-1}\bmu\\[1ex]
-\smhalf\vecof(\bSigma^{-1})
\end{array}
\right]
\quad
\mbox{and}
\quad
\bdetavvech\equiv
\left[     
\begin{array}{c}
\bSigma^{-1}\bmu\\[1ex]
-\smhalf\bD_d^T\vecof(\bSigma^{-1})
\end{array}
\right].
$$
The two natural parameter vectors can be mapped between each other using
\begin{equation}
\bdetavvech=\mbox{blockdiag}(\bI_d,\bD_d^T)\bdetavvec
\quad\mbox{and}\quad
\bdetavvec=\mbox{blockdiag}(\bI_d,\bD_d^{+T})\bdetavvech.
\label{eq:vecvechmapsMVN}
\end{equation}

Now we explain the interplay between the ``vec'' and ``vech'' forms of the
Inverse G-Wishart distribution. Let the $d\times d$ matrix $\bV$
have an $\mbox{Inverse-G-Wishart}(G,\xi,\bLambda)$ distribution.
Then the density function of $\bV$ is 
$$\pDens(\bV)\propto
\exp\left\{\left[
\begin{array}{c}
\log|\bV|\\[1ex]
\vecof(\bV^{-1})
\end{array}
\right]^T\bdetaVvec
\right\}
=\exp\left\{\left[
\begin{array}{c}
\log|\bV|\\[1ex]
\vech(\bV^{-1})
\end{array}
\right]^T\bdetaVvech
\right\}
$$
where
$$\bdetaVvec\equiv
\left[
\begin{array}{c}
-\smhalf(\xi+1)\\[1ex]
-\smhalf\vecof(\bLambda)
\end{array}
\right]
\quad\mbox{and}\quad
\bdetaVvech\equiv
\left[
\begin{array}{c}
-\smhalf(\xi+1)\\[1ex]
-\smhalf\bD_d^T\vecof(\bLambda)
\end{array}
\right].
$$
Mappings between the two natural parameter vectors are as follows:
\begin{equation}
\bdetaVvech=\mbox{blockdiag}(1,\bD_d^T)\,\bdetaVvec
\quad\mbox{and}\quad
\bdetaVvec=\mbox{blockdiag}(1,\bD_d^{+T})\,\bdetaVvech.
\label{eq:vecvechmapsIGW}
\end{equation}

\section{Justification of Algorithm \ref{alg:IterInvGWishFrag}}\label{sec:mainAlgDrv}

We now provide justification for Algorithm \ref{alg:IterInvGWishFrag}, which is
concerned with the graph and natural parameter updates for the iterated
Inverse G-Wishart fragment.

\subsection{The Updates for $\mSUBpSigmaATOSigma$}\label{sec:drvFirstMsg}
As a function of $\bSigma$, 
$$\log\,\pDens(\bSigma|\bA)=
\left[
\begin{array}{c}
\log|\bSigma|\\[1ex]
\vech(\bSigma^{-1})
\end{array}
\right]^T
\left[
\begin{array}{c}
-\smhalf\,(\xi+2)\\[1ex]
-\smhalf\bD_d^T\vecof(\bA^{-1})
\end{array}
\right]+\mbox{const}
$$
where `const' denotes terms that do not depend on $\bSigma$.
Hence 
$$\mSUBpSigmaATOSigma=\exp\left\{ 
\left[
\begin{array}{c}
\log|\bSigma|\\[1ex]
\vech(\bSigma^{-1})
\end{array}
\right]^T\etaSUBpSigmaATOSigma\right\}
$$
where
\begin{equation}
\etaSUBpSigmaATOSigma
=
\left[
\begin{array}{c}
-\smhalf\,(\xi+2)\\[1ex]
-\smhalf\bD_d^T\vecof\Big(E_{\qDens}(\bA^{-1})\Big)
\end{array}
\right]
\label{eq:roosterCrow}
\end{equation}
and $E_{\qDens}$ denotes expectation with respect to the normalization of
$$\mSUBpSigmaATOA\,\mSUBATOpSigmaA.$$
Let $\qDens(\bA)$ denote this normalized density function. Then
$\qDens(\bA)$ is an Inverse-G-Wishart distribution with graph 
$\GSUBpSigmaATOA\in\{\Gfull,\Gdiag\}$ and natural parameter 
vector $\etaSUBpSigmaACONNA$. From Result \ref{eq:recipMoment}, 
$$E_{\qDens}(\bA^{-1})
=\left\{
\begin{array}{l}
\Big\{\big(\etaSUBpSigmaACONNA\big)_1+\smhalf(d+1)\Big\}
\left\{\vecof^{-1}\Big(\bD_d^{+T}\big(\etaSUBpSigmaACONNA\big)_2\Big)\right\}^{-1}\\[2ex]
\qquad\qquad\qquad\qquad\qquad\qquad\qquad\qquad\qquad\qquad\mbox{if $\GSUBpSigmaATOA=\Gfull$},\\[2ex] 
\Big\{\big(\etaSUBpSigmaACONNA\big)_1+1\Big\}
\left\{\vecof^{-1}\Big(\bD_d^{+T}\big(\etaSUBpSigmaACONNA\big)_2\Big)\right\}^{-1}\\[2ex] 
\qquad\qquad\qquad\qquad\qquad\qquad\qquad\qquad\qquad\qquad
\mbox{if $\GSUBpSigmaATOA=\Gdiag$}.
\end{array}
\right.
$$
Noting that the first factor of $E_{\qDens}(\bA^{-1})$ is
$(\etaSUBpSigmaACONNA)_1+\omega_1$,
where
$$\omega_1=\omega_1(d,G)=\left\{
\begin{array}{ll}
(d+1)/2    &  \mbox{if $G=\Gfull$}\\[1ex]
1          &  \mbox{if $G=\Gdiag$},
\end{array}
\right.
$$
the first update of $E_{\qDens}(\bA^{-1})$ in Algorithm \ref{alg:IterInvGWishFrag} is justified.
Lastly, we need to possibly adjust for the fact that $\mSUBpSigmaATOSigma$
is proportional to an Inverse G-Wishart density function with $G=\Gdiag$. 
This is achieved by the conditional step:

\centerline{If $\GSUBpSigmaATOSigma=\Gdiag$ then 
$E_{\qDens}(\bA^{-1})\thickarrow\diag\left\{\diagonal\Big(E_{\qDens}(\bA^{-1})\Big)\right\}.$}

\subsection{The Updates for $\mSUBpSigmaATOA$}

As a function of $\bA$, 
$$\log\,\pDens(\bSigma|\bA)=
\left[
\begin{array}{c}
\log|\bA|\\[1ex]
\vech(\bA^{-1})
\end{array}
\right]^T
\left[
\begin{array}{c}
-(\xi+2-2\omega_2)/2\\[1ex]
-\smhalf\bD_d^T\vecof(\bSigma^{-1})
\end{array}
\right]+\mbox{const}
$$
where
\begin{equation}
\omega_2=\omega_2(d,G)=\left\{
\begin{array}{ll}
(d+1)/2    &  \mbox{if $G=\Gfull$},\\[1ex]
1          &  \mbox{if $G=\Gdiag$}
\end{array}
\right.
\label{eq:fryingPan}
\end{equation}
and `const' denotes terms that do not depend on $\bA$.
Hence 
$$\mSUBpSigmaATOA=\exp\left\{ 
\left[
\begin{array}{c}
\log|\bA|\\[1ex]
\vech(\bA^{-1})
\end{array}
\right]^T\etaSUBpSigmaATOA\right\}
$$
where
\begin{equation}
\etaSUBpSigmaATOA
=
\left[
\begin{array}{c}
-(\xi+2-2\omega_2)/2\\[1ex]
-\smhalf\bD_d^T\vecof\Big(E_{\qDens}(\bSigma^{-1})\Big)
\end{array}
\right]
\label{eq:boobyTrap}
\end{equation}
and $E_{\qDens}$ denotes expectation with respect to the normalization of
$$\mSUBpSigmaATOSigma\,\mSUBSigmaTOpSigmaA.$$
Let $\qDens(\bSigma)$ denote this normalized density function.
Then $\qDens(\bSigma)$ is an Inverse-G-Wishart distribution with 
graph $\GSUBpSigmaATOSigma\in\{\Gfull,\Gdiag\}$
and natural parameter vector $\etaSUBpSigmaACONNSigma$.
From Result \ref{eq:recipMoment},
$$E_{\qDens}(\bSigma^{-1})
=\left\{
\begin{array}{l}
\Big\{\big(\etaSUBpSigmaACONNSigma\big)_1+\smhalf(d+1)\Big\}
\left\{\vecof^{-1}\Big(\bD_d^{+T}\big(\etaSUBpSigmaACONNSigma\big)_2\Big)\right\}^{-1}\\[2ex]
\qquad\qquad\qquad\qquad\qquad\qquad\qquad\qquad\qquad\qquad\mbox{if $\GSUBpSigmaATOSigma=\Gfull$},\\[2ex] 
\Big\{\big(\etaSUBpSigmaACONNSigma\big)_1+1\Big\}
\left\{\vecof^{-1}\Big(\bD_d^{+T}\big(\etaSUBpSigmaACONNSigma\big)_2\Big)\right\}^{-1}\\[2ex] 
\qquad\qquad\qquad\qquad\qquad\qquad\qquad\qquad\qquad\qquad
\mbox{if $\GSUBpSigmaATOSigma=\Gdiag$}.
\end{array}
\right.
$$
Noting that the first factor of $E_{\qDens}(\bSigma^{-1})$ is
$(\etaSUBpSigmaACONNSigma)_1+\omega_2$,
where $\omega_2$ is given by (\ref{eq:fryingPan}),
the first update of $E_{\qDens}(\bSigma^{-1})$ in Algorithm \ref{alg:IterInvGWishFrag} is justified.
Finally, there is the possible need to adjust for the fact that $\mSUBpSigmaATOA$
is proportional to an Inverse G-Wishart density function with $G=\Gdiag$. 
This is achieved by the conditional step:

\centerline{If $\GSUBpSigmaATOA=\Gdiag$ then 
$E_{\qDens}(\bSigma^{-1})\thickarrow\diag\left\{\diagonal\Big(E_{\qDens}(\bSigma^{-1})\Big)\right\}.$}

\section{Illustrative Example Variational Message Passing Details}

The variational message passing approach to fitting and approximate
inference for statistical models is still quite a new concept.
In this section we provide details on the approach for the 
illustrative example involving  the $t$ response linear mixed model
described in Section \ref{sec:illustrative}.

\subsection{Data and Hyperparameter Inputs}

Let $\by$ be the vector of responses as defined in (\ref{eq:yXZdefn}).
Also, let 
$$\bC=[\bX\ \bZ]$$
be the full design matrix, where the matrices $\bX$ and $\bZ$
are as defined in (\ref{eq:yXZdefn}).
The data inputs are $\by$ and $\bC$.

The hyperparameter inputs are
$$\sigma_{\bbeta},\ssigsq,\lambda_{\nu},\sSigmaOne,\ldots\sSigmaq>0.$$
		
\subsection{Factor to Stochastic Node Parameter Initialisations}

\noindent
Initialize $\GSUBpATOA$ and  $\etaSUBpATOA$ via a call to Algorithm \ref{alg:InvGWishPriorFrag} with
hyperparameter inputs:
$$G_{\mbox{\tiny{$\bTheta$}}}=\Gdiag,\quad\xiTheta=1\quad\mbox{and}\quad 
\LambdaTheta=\{2\diag(\sSigmaOne^2,\ldots\sSigmaq^2)\}^{-1}.
$$ 
\vskip5mm
\noindent
Initialize $\GSUBpaTOa$ and $\etaSUBpaTOa$ via a call to Algorithm \ref{alg:InvGWishPriorFrag} with
hyperparameter inputs:
$$G_{\mbox{\tiny{$\bTheta$}}}=\Gdiag,\quad\xiTheta=1\quad\mbox{and}\quad 
\LambdaTheta=(\ssigma^2)^{-1}.
$$ 

\noindent
Note that the initialisations of $\GSUBpATOA$, $\etaSUBpATOA$, $\GSUBpaTOa$ and $\etaSUBpaTOa$ 
are part of the prior impositions for $\bSigma$ and $\sigma^2$. These four 
factor to stochastic node parameters remain constant throughout the variational
message passing iterations.

\vskip5mm
\noindent
Initialize 
$$
\etaSUBpupsilonTOupsilon\longleftarrow
\left[
\begin{array}{c}
0\\[1ex]
-\lambda_{\nu}
\end{array}
\right].
$$

\noindent
This initialization of $\etaSUBpupsilonTOupsilon$ corresponds to the prior imposition
for $\upsilon$. This factor to stochastic node natural parameter remains constant
throughout the variational message passing iterations.

\vskip5mm
The remaining factor to stochastic node natural parameters in the 
Figure \ref{fig:tMixModFacGraph} factor graph are updated in the
variational message passing iterations, but require initial values. In theory,
they can be set to any legal value according to the relevant exponential family.
The following initialisations, which are used in the code that
produced Figure \ref{fig:tMixModPosterDens}, are simple legal natural 
parameter vectors:

$$\GSUBpSigmaATOA\longleftarrow\Gdiag,\ \ 
\etaSUBpSigmaATOA\longleftarrow\left[    
\begin{array}{c}
-\frac{1}{2}\\[1ex]
-\frac{1}{2}\bD_q^T\vecof(\bI_q)
\end{array}
\right],
$$

$$\GSUBpSigmaATOSigma\longleftarrow\Gfull,\ \  
\etaSUBpSigmaATOSigma\longleftarrow\left[
\begin{array}{c}
-\frac{1}{2}\\[1ex]
-\frac{1}{2}\bD_q^T\vecof(\bI_q)
\end{array}
\right],
$$

$$\GSUBpsigsqaTOa\longleftarrow\Gdiag,\ \ 
\etaSUBpsigsqaTOa\longleftarrow\left[   
\begin{array}{c}
-2\\[1ex]
-1
\end{array}
\right],
$$

$$\GSUBpsigsqaTOsigsq\longleftarrow\Gfull,\
\etaSUBpsigsqaTOsigsq\longleftarrow\left[   
\begin{array}{c}
-2\\[1ex]
-1
\end{array}
\right],
$$

$$\etaSUBpbetauSigmaTOSigma\longleftarrow\left[    
\begin{array}{c}
-\frac{1}{2}\\[1ex]
-\frac{1}{2}\vecof(\bI_q)
\end{array}
\right],\quad
\etaSUBpbetauSigmaTObetau\longleftarrow\left[   
\begin{array}{c}
\bzero_{p+m q}\\[1ex]
-\frac{1}{2}\vecof(\bI_{p+m q})
\end{array}
\right],
$$ 

$$\etaSUBpybetausigsqbTObetau\longleftarrow\left[    
\begin{array}{c}
\bzero_{p+m q}\\[1ex]
-\frac{1}{2}\vecof(\bI_{p+mq})
\end{array}
\right],\quad 
\etaSUBpybetausigsqbTOsigsq
\longleftarrow
\left[   
\begin{array}{c}
-2\\[1ex]
-1
\end{array}
\right]
$$
and
$$\etaSUBpbupsilonTOupsilon\longleftarrow\left[               
\begin{array}{c}
1\\[1ex]
-1.1
\end{array}
\right].
$$
The messages involving the $b_{\ell}$, $1\le\ell\le N$, nodes do not need to 
be included here since there messages are subsumed
in the calculations used for the natural parameter
updates for the model parameters in Algorithm 2 of
McLean \myand Wand (2019).

\subsection{Variational Message Passing Iterations}

With all factor to stochastic node initialisations accomplished, now
we describe the iterative updates inside the variational message
passing cycle loop. Each iteration involves:
\begin{itemize}
\item updating the stochastic node to factor message parameters.
\item updating the factor to stochastic node message parameters.
\end{itemize}

\subsubsection{Stochastic Node to Factor Message Parameter Updates}

The stochastic node to factor message updates are quite simple
and follow from, e.g., equation (7) of Wand (2017).
For the Figure \ref{fig:tMixModFacGraph} factor graph the updates are:

$$\GSUBATOpSigmaA\longleftarrow\GSUBpATOA,\quad
\etaSUBATOpSigmaA\longleftarrow\etaSUBpATOA,
$$
$$
\GSUBSigmaTOpSigmaA\longleftarrow\Gfull,\quad
\etaSUBSigmaTOpSigmaA\longleftarrow\etaSUBpbetauSigmaTOSigma,
$$
$$
\etaSUBSigmaTOpbetauSigma\longleftarrow\etaSUBpSigmaATOSigma,\quad
\etaSUBbetauTOpbetauSigma\longleftarrow\etaSUBpybetausigsqbTObetau,
$$
$$\etaSUBbetauTOpybetausigsqb\longleftarrow\etaSUBpbetauSigmaTObetau,\quad
\GSUBaTOpsigsqa\longleftarrow\GSUBpaTOa,
$$
$$
\etaSUBaTOpsigsqa\longleftarrow\etaSUBpaTOa,\quad
\GSUBsigsqTOpsigsqa\longleftarrow\Gfull,
$$
$$\etaSUBsigsqTOpsigsqa\longleftarrow\etaSUBpybetausigsqbTOsigsq,\quad
\etaSUBsigsqTOpybetausigsqb\longleftarrow\etaSUBpsigsqaTOsigsq$$
and
$$\etaSUBupsilonTOpbupsilon\longleftarrow\etaSUBpupsilonTOupsilon.$$

\vskip5mm
Some additional remarks concerning stochastic node to factor updates
are:
\begin{itemize}
\item The stochastic node to factor messages corresponding to the extremities of the
Figure \ref{fig:tMixModFacGraph} factor graph, such as the message from $\bA$ to $\pDens(\bA)$, are not 
required in the variational message passing iterations. Therefore, updates for
these messages can be omitted.
\item Some of the stochastic node to factor message parameter updates, such as that for 
$\etaSUBATOpSigmaA$, remain constant throughout the iterations. However, for
simplicity of exposition, we list all of the updates together.
\end{itemize}

\subsubsection{Factor to Stochastic Node Message Parameter Updates}

The updates for the parameters of factor to stochastic node messages
are a good deal more complicated than the reverse messages. 
For the illustrative example, these updates are encapsulated in
three algorithms across three different articles. Algorithm \ref{alg:IterInvGWishFrag} 
plays an important role for the variance and covariance matrix parameter
parts of the factor graph.

\begin{itemize}
\setlength\itemsep{0pt}
\item[] \textrm{Use Algorithm \ref{alg:IterInvGWishFrag} with:}
\begin{itemize}
\setlength\itemsep{0pt}
\item[] \textbf{\small Shape Parameter Input}: $1$.
\item[] \textbf{\small Graph Inputs}:  $\GSUBsigsqTOpsigsqa$, $\GSUBaTOpsigsqa$.
\item[] \textbf{\small Natural Parameter Inputs:} $\etaSUBsigsqTOpsigsqa$, $\etaSUBpsigsqaTOsigsq$ 
 $\etaSUBaTOpsigsqa$, $\etaSUBpsigsqaTOa$
 \item[] \textbf{\small Outputs:} 
        $\GSUBpsigsqaTOsigsq$, $\etaSUBpsigsqaTOsigsq$, 
	$\GSUBpsigsqaTOa$, $\etaSUBpsigsqaTOa$
\end{itemize}

\item[] \textrm{Use Algorithm \ref{alg:IterInvGWishFrag} with:}
\begin{itemize}
\setlength\itemsep{0pt}
\item[] \textbf{\small Shape Parameter Input}: $2q$
\item[] \textbf{\small Graph Inputs}: $\GSUBSigmaTOpSigmaA$, $\GSUBATOpSigmaA$
\item[] \textbf{\small Natural Parameter Inputs:} $\etaSUBSigmaTOpSigmaA$, $\etaSUBpSigmaATOSigma$, 
$\etaSUBATOpSigmaA$, $\etaSUBpSigmaATOA$
\item[] \textbf{\small Outputs:} $\GSUBpSigmaATOSigma$, $\etaSUBpSigmaATOSigma$,
$\GSUBpSigmaATOA$, $\etaSUBpSigmaATOA$
\end{itemize}
\item[] \textrm{Use the Gaussian Penalisation Fragment of Wand (2017, Section 4.1.4):}
\begin{itemize}
\setlength\itemsep{0pt}
\item[] \textbf{\small Hyperparameter Input:} $\sigma^2_{\bbeta}$
\item[] \textbf{\small Natural Parameter Inputs:} $\etaSUBbetauTOpbetauSigma$, $\etaSUBpbetauSigmaTObetau$,\\
\null$\qquad\qquad\qquad\qquad\qquad\quad\,\etaSUBSigmaTOpbetauSigma$, $\etaSUBpbetauSigmaTOSigma$
\item[] \textbf{\small Outputs:} $\etaSUBpbetauSigmaTObetau$, $\etaSUBpbetauSigmaTOSigma$
\end{itemize}
\item[] \textrm{Use the $t$ Likelihood Fragment of McLean \myand Wand (2019, Algorithm 2):}
\begin{itemize}
\setlength\itemsep{0pt}
\item[] \textbf{\small Data Inputs:} $\by$, $\bC$
\item[] \textbf{\small Natural Parameter Inputs:} $\etaSUBbetauTOpybetausigsqb$, $\etaSUBpybetausigsqbTObetau$,\\
\null$\qquad\qquad\qquad\qquad\qquad\quad\,\etaSUBsigsqTOpybetausigsqb$, $\etaSUBpybetausigsqbTOsigsq$,\\
\null$\qquad\qquad\qquad\qquad\qquad\quad\,\etaSUBupsilonTOpbupsilon$, $\etaSUBpbupsilonTOupsilon$
\item[] \textbf{\small Outputs:} $\etaSUBpybetausigsqbTObetau$, $\etaSUBpybetausigsqbTOsigsq$,
$\etaSUBpbupsilonTOupsilon$
\end{itemize}
\end{itemize}

Regarding, the last two fragment updates it should be noted
that Wand (2017) and McLean \myand Wand (2019) work with
the ``$\vecof$'' versions of Multivariate Normal and Inverse G-Wishart
natural parameter vectors. To match the ``$\vech$'' natural parameter 
forms used in Algorithms \ref{alg:InvGWishPriorFrag} and 
\ref{alg:IterInvGWishFrag} of the current article conversions
given by (\ref{eq:vecvechmapsMVN}) and (\ref{eq:vecvechmapsIGW})
are required.

\subsection{Determination of Posterior Density Function Approximations}

After convergence of the variational message passing iterations, 
the optimal $\qDens^*$-densities for each stochastic node are obtained by 
multiplying each of the messages that pass messages to that node.
See, for example, (10) of Wand (2017). We now give
details for the model parameters $\bSigma$, $\sigma^2$, $(\bbeta,\bu)$ and
$\upsilon$.

\subsubsection{Determination of $\qDens^*(\bSigma)$}

From (10) of Wand (2017):
$$\qDens^*(\bSigma)\propto\exp\left\{
\left[
\begin{array}{c}
\log|\bSigma|\\[1ex]
\vech(\bSigma^{-1})
\end{array}
\right]^T
\left(\etaSUBpSigmaATOSigma+\etaSUBpbetauSigmaTOSigma\right)
\right\}.
$$
It is apparent that $\qDens^*(\bSigma)$ is an Inverse Wishart 
density function with natural parameter vector
$$\biggerbdeta_{\qDens(\bSigma)}\equiv \etaSUBpSigmaATOSigma+\etaSUBpbetauSigmaTOSigma.$$

\subsubsection{Determination of $\qDens^*(\sigma^2)$}

Using (10) of Wand (2017):
$$\qDens^*(\sigma^2)\propto\exp\left\{
\left[
\begin{array}{c}
\log(\sigma^2)\\[1ex]
1/\sigma^2
\end{array}
\right]^T
\left(\etaSUBpsigsqaTOsigsq+\etaSUBpybetausigsqbTOsigsq\right)
\right\}.
$$
We see that $\qDens^*(\sigma^2)$ is an Inverse Chi-Squared
density function with natural parameter vector
$$\biggerbdeta_{\qDens(\sigma^2)}\equiv \etaSUBpsigsqaTOsigsq+\etaSUBpybetausigsqbTOsigsq.$$

\subsubsection{Determination of $\qDens^*(\bbeta,\bu)$}

Another application of (10) of Wand (2017) leads to:
$$\qDens^*(\bbeta,\bu)\propto\exp\left\{
\left[
\begin{array}{c}
\bbeta\\[0ex]
\bu\\[1ex]
\vech\left(\left[     
\begin{array}{c}
\bbeta\\[0ex]
\bu
\end{array}
\right]
\left[     
\begin{array}{c}
\bbeta\\[0ex]
\bu
\end{array}
\right]^T
\right)
\end{array}
\right]^T
\left(\etaSUBpbetauSigmaTObetau+\etaSUBpybetausigsqbTObetau\right)
\right\}.
$$
We then have $\qDens^*(\bbeta,\bu)$ having a Multivariate Normal
density function with natural parameter vector
$$\biggerbdeta_{\qDens(\bbeta,\bu)}\equiv \etaSUBpbetauSigmaTObetau+\etaSUBpybetausigsqbTObetau.$$

\subsubsection{Determination of $\qDens^*(\upsilon)$}

One last application of (10) of Wand (2017) gives:
$$\qDens^*(\upsilon)\propto\exp\left\{
\left[
\begin{array}{c}
\upsilon\log(\upsilon)-\log\{\Gamma(\upsilon)\}\\[1ex]
\upsilon
\end{array}
\right]^T
\left(\etaSUBpbupsilonTOupsilon+\etaSUBpupsilonTOupsilon\right)
\right\}.
$$
Therefore, $\qDens^*(\upsilon)$ is a Moon Rock density function with natural parameter
vector
$$\biggerbdeta_{\qDens(\upsilon)}\equiv\etaSUBpbupsilonTOupsilon+\etaSUBpupsilonTOupsilon.$$

\subsection{Conversion from Natural Parameters to Common Parameters}

A final set of steps involves conversion of the $\qDens^*$-densities to common parameter
forms. 

\subsubsection{Conversion of $\qDens^*(\bSigma)$ to Common Parameter Form}

The common parameter form of $\qDens^*(\bSigma)$ is the 
$\mbox{Inverse-G-Wishart}(\Gfull,\xi_{\qDens(\bSigma)},\bLambda_{\qDens(\bSigma)})$
density function where
$$\xi_{\qDens(\bSigma)}=-2\big(\biggerbdeta_{\qDens(\bSigma)}\big)_1-2
\quad\mbox{and}\quad
\bLambda_{\qDens(\bSigma)}=-2\vecof^{-1}\Big(\bD_q^{+T}\big(\biggerbdeta_{\qDens(\bSigma)}\big)_2\Big).
$$
Alternatively, $\qDens^*(\bSigma)$ is the 
$\mbox{Inverse-Wishart}(\kappa_{\qDens(\bSigma)},\bLambda_{\qDens(\bSigma)})$ density
function, as defined by (\ref{eq:GelmanTable}), where
$$\kappa_{\qDens(\bSigma)}=\xi_{\qDens(\bSigma)}-q+1.$$

\subsubsection{Conversion of $\qDens^*(\sigma^2)$ to Common Parameter Form}

The common parameter form of  $\qDens^*(\sigma^2)$ is the 
$\mbox{{\rm Inverse}-$\chi^2$}(\delta_{\qDens(\sigma^2)},\lambda_{\qDens(\sigma^2)})$
density function where
$$\delta_{\qDens(\sigma^2)}=-2\big(\biggerbdeta_{\qDens(\sigma^2)}\big)_1-2
\quad\mbox{and}\quad
\lambda_{\qDens(\sigma^2)}=-2\big(\biggerbdeta_{\qDens(\sigma^2)}\big)_2.
$$

\subsubsection{Conversion of $\qDens^*(\bbeta,\bu)$ to Common Parameter Form}

The common parameter form of $\qDens^*(\bbeta,\bu)$ is the 
$N\big(\bmu_{\qDens(\bbeta,\bu)},\bSigma_{\qDens(\bbeta,\bu)}\big)$ density function
where
$$\bmu_{\qDens(\bbeta,\bu)}=-\smhalf\left\{\vecof^{-1}
\Big(\bD_{p+mq}^{+T} \big(\biggerbdeta_{\qDens(\bbeta,\bu)}\big)_2\Big)
\right\}^{-1}\big(\biggerbdeta_{\qDens(\bbeta,\bu)}\big)_1
$$
and
$$
\bSigma_{\qDens(\bbeta,\bu)}=-\smhalf\left\{\vecof^{-1}
\Big(\bD_{p+mq}^{+T} \big(\biggerbdeta_{\qDens(\bbeta,\bu)}\big)_2\Big)
\right\}^{-1}.
$$
Here $\big(\biggerbdeta_{\qDens(\bbeta,\bu)}\big)_1$ denotes the first $p+mq$ entries of
$\biggerbdeta_{\qDens(\bbeta,\bu)}$ and $\big(\biggerbdeta_{\qDens(\bbeta,\bu)}\big)_2$
denotes the remaining entries of the same vector.

\subsubsection{Conversion of $\qDens^*(\upsilon)$ to Common Parameter Form and Conversion to $\qDens^*(\nu)$ }

Recall that $\qDens^*(\upsilon)$ is a Moon Rock density function. The Moon Rock distribution
is not as established as the other distributions appearing in this subsection. Nevertheless,
the web-supplement of McLean \myand Wand (2019) defines a random variable $x$ to have
a Moon Rock distribution with parameters $\alpha>0$ and $\beta>\alpha$, written
$x\sim\mbox{Moon-Rock}(\alpha,\beta)$, if the density function of $x$ is 
$$\pDens(x)=\left[\int_0^{\infty}\{t^t/\Gamma(t)\}^{\alpha}\exp(-\beta\,t)\,dt\right]^{-1}
\{x^x/\Gamma(x)\}^{\alpha}\exp(-\beta\,x),\quad x>0.
$$
Therefore, $\qDens^*(\upsilon)$ has a $\mbox{Moon-Rock}(\alpha_{\qDens(\upsilon)},\beta_{\qDens(\upsilon)})$
density function where
$$\alpha_{\qDens(\upsilon)}=\big(\biggerbdeta_{\qDens(\upsilon)}\big)_1
\quad\mbox{and}\quad\beta_{\qDens(\upsilon)}=\,-\big(\biggerbdeta_{\qDens(\upsilon)}\big)_2.
$$
Explicitly,
$$\qDens^*(\upsilon)=\left[\int_0^{\infty}\{t^t/\Gamma(t)\}^{\alpha_{\qDens(\upsilon)}}
\exp(-\beta_{\qDens(\upsilon)}\,t)\,dt\right]^{-1}
\{\upsilon^{\upsilon}/\Gamma(\upsilon)\}^{\alpha_{\qDens(\upsilon)}}
\exp(-\beta_{\qDens(\upsilon)}\,\upsilon),\quad \upsilon>0.
$$
Lastly, we note that since $\nu=2\upsilon$ the $\qDens^*$-density function of 
$\nu$ is 
{\setlength\arraycolsep{3pt}
\begin{eqnarray*}
&&\qDens^*(\nu)=\smhalf
\left[\int_0^{\infty}\{t^t/\Gamma(t)\}^{\alpha_{\qDens(\upsilon)}}
\exp(-\beta_{\qDens(\upsilon)}\,t)\,dt\right]^{-1}\\[1ex]
&&\qquad\qquad\qquad\qquad\times
\{(\nu/2)^{\nu/2}/\Gamma(\nu/2)\}^{\alpha_{\qDens(\upsilon)}}
\exp\big(-\smhalf \beta_{\qDens(\upsilon)}\nu\big),
\quad\nu>0.
\end{eqnarray*}
}

\section*{Reference}

\bib
Novomestky, F. (2012). 
\textsf{matrixcalc}: Collection of functions for matrix calculations.
\textsf{R} package. \texttt{https://CRAN.R-project.org/package=matrixcalc}

\end{document}

%% file: igwmacros.tex
\def\etaSUBpThetaOneThetaTwoTOThetaTwo{
\biggerbdeta_{\mbox{\footnotesize$p(\bTheta_1|\bTheta_2)\to\bTheta_2$}}}
\def\GSUBpsigsqaTOsigsq{G_{\mbox{\footnotesize$\pDens(\sigma^2|\,a)\rightarrow\sigma^2$}}}
\def\GSUBaTOpsigsqa{G_{\mbox{\footnotesize$a\rightarrow\pDens(\sigma^2|\,a)$}}}
\def\GSUBpsigsqaTOa{G_{\mbox{\footnotesize$\pDens(\sigma^2|\,a)\rightarrow a$}}}
\def\xiTheta{\xi_{\mbox{\tiny{$\bTheta$}}}}
\def\LambdaTheta{\bLambda_{\mbox{\tiny{$\bTheta$}}}}
\def\GSUBpATOA{G_{\mbox{\footnotesize$\pDens(\bA)\rightarrow\bA$}}}
\def\etaSUBpATOA{\biggerbdeta_{\mbox{\footnotesize$\pDens(\bA)\rightarrow \bA$}}}
\def\GSUBpaTOa{G_{\mbox{\footnotesize$\pDens(a)\rightarrow a$}}}
\def\etaSUBpaTOa{\biggerbdeta_{\mbox{\footnotesize$\pDens(a)\rightarrow a$}}}

\def\etaSUBpupsilonTOupsilon{\biggerbdeta_{\mbox{\footnotesize$\pDens(\upsilon)\rightarrow\upsilon$}}}

\def\etaSUBpbupsilonTOupsilon{\biggerbdeta_{\mbox{\footnotesize$\pDens(\bb|\,\upsilon)\rightarrow\upsilon$}}}
\def\etaSUBpybetausigsqbTOsigsq
{\biggerbdeta_{\mbox{\footnotesize$\pDens(\by|\,\bbeta,\bu,\sigma^2,\bb)\rightarrow\sigma^2$}}}
\def\etaSUBpybetausigsqbTObetau
{\biggerbdeta_{\mbox{\footnotesize$\pDens(\by|\,\bbeta,\bu,\sigma^2,\bb)\rightarrow(\bbeta,\bu)$}}}
\def\etaSUBupsilonTOpbupsilon
{\biggerbdeta_{\mbox{\footnotesize$\upsilon\rightarrow\pDens(\bb|\,\upsilon)$}}}

\def\etaSUBbetauTOpybetausigsqb
{\biggerbdeta_{\mbox{\footnotesize$(\bbeta,\bu)\rightarrow\pDens(\by|\,\bbeta,\bu,\sigma^2,\bb)$}}}
\def\etaSUBsigsqTOpybetausigsqb
{\biggerbdeta_{\mbox{\footnotesize$\sigma^2\rightarrow\pDens(\by|\,\bbeta,\bu,\sigma^2,\bb)$}}}
\def\kappaSubSigma{\kappa_{\scriptscriptstyle\bSigma}}
\def\LambdaSubSigma{\bLambda_{\scriptscriptstyle\bSigma}}
\def\LambdaSubTheta{\bLambda_{\scriptscriptstyle\bTheta}}
\def\GsubTheta{G_{\scriptscriptstyle\bTheta}}
\def\xiSubTheta{\xi_{\scriptscriptstyle\bTheta}}
\def\myand{\&\ }
\def\deltaMP{\delta_{\mbox{\tiny MP}}}
\def\nuMP{\nu_{\mbox{\tiny MP}}}
\def\nuHW{\nu_{\mbox{\tiny HW}}}
\def\bBMP{\bB_{\mbox{\tiny MP}}}
\def\bX{\boldsymbol{X}}
\def\bM{\boldsymbol{M}}
\def\Psc{{\mathcal P}}
\def\Dsc{{\mathcal D}}
\def\simind{\stackrel{{\tiny \mbox{ind.}}}{\sim}}
\def\bv{\boldsymbol{v}}
\def\diag{\mbox{diag}}
\def\vecof{\mbox{\rm vec}}
\def\vech{\mbox{\rm vech}}
\def\naturalNumbers{{\mathbb N}}
\def\ba{\boldsymbol{a}}
\def\bb{\boldsymbol{b}}

\def\bu{\boldsymbol{u}}
\def\bv{\boldsymbol{v}}

\def\by{\boldsymbol{y}}
\def\bA{\boldsymbol{A}}
\def\bB{\boldsymbol{B}}
\def\bC{\boldsymbol{C}}
\def\bD{\boldsymbol{D}}
\def\bLambda{\boldsymbol{\Lambda}}
\def\smhalf{{\textstyle{\frac{1}{2}}}}
\def\diagonal{\mbox{diagonal}}
\def\bT{\boldsymbol{T}}

\def\bV{\boldsymbol{V}}

\def\bX{\boldsymbol{X}}
\def\bY{\boldsymbol{Y}}
\def\bZ{\boldsymbol{Z}}
\def\bdeta{\boldsymbol{\eta}}
\def\bSigma{\boldsymbol{\Sigma}}
\def\bTheta{\boldsymbol{\Theta}}
\def\tr{\mbox{tr}}
\def\thickarrow{\longleftarrow}
\def\jump{\vskip3mm\noindent}

\def\blockdiagdum{\mathop{\mbox{\rm blockdiag}}}
\def\blockdiag#1{\blockdiagdum_{#1}}
\def\bbeta{\boldsymbol{\beta}}
\def\bzero{\boldsymbol{0}}
\def\bI{\boldsymbol{I}}
\def\bmu{\boldsymbol{\mu}}
\def\btheta{\boldsymbol{\theta}}

\def\bib{\vskip12pt\par\noindent\hangindent=1 true cm\hangafter=1}
\def\bigtimes{{\Large\mbox{$\times$}}}
\def\padzero{{\large\mbox{\ $0$\ }}}

\def\ssigsq{s_{\sigma^2}}

\def\sSigmaOne{s_{\mbox{\tiny{$\bSigma,1$}}}}
\def\sSigmaTwo{s_{\mbox{\tiny{$\bSigma,2$}}}}
\def\sSigmaq{s_{\mbox{\tiny{$\bSigma,q$}}}}
\def\sSigmad{s_{\mbox{\tiny{$\bSigma,d$}}}}
\def\pDens{p}
\def\qDens{q}
\def\bdetavvech{\bdeta_v^{\mbox{\tiny vech}}}
\def\bdetavvec{\bdeta_v^{\mbox{\tiny vec}}}
\def\bdetaVvech{\bdeta_{\mbox{\tiny$\bV$}}^{\mbox{\tiny vech}}}
\def\bdetaVvec{\bdeta_{\mbox{\tiny$\bV$}}^{\mbox{\tiny vec}}}
\def\bbetaTrue{\bbeta_{\mbox{\tiny true}}}
\def\sigsqTrue{\sigma^2_{\mbox{\tiny true}}}
\def\bSigmaTrue{\bSigma_{\mbox{\tiny true}}}
\def\nuTrue{\nu_{\mbox{\tiny true}}}
\def\halfnu{\upsilon}
\def\biggerbdeta{\mbox{\Large $\bdeta$}}
\def\biggerm{\mbox{\Large $m$}}
\def\ssigma{s_{\sigma}}
\def\nusigma{\nu_{\sigma}}
\def\alphasigsq{\alpha_{\sigma^2}}
\def\betasigsq{\beta_{\sigma^2}}
\def\etaSUBpSigmaATOSigma{\biggerbdeta_{\mbox{\scriptsize$\pDens(\bSigma|\bA)\rightarrow\bSigma$}}}
\def\etaSUBpSigmaATOA{\biggerbdeta_{\mbox{\scriptsize$\pDens(\bSigma|\bA)\rightarrow\bA$}}}
\def\etaSUBSigmaTOpSigmaA{\biggerbdeta_{\mbox{\scriptsize$\bSigma\rightarrow \pDens(\bSigma|\bA)$}}}
\def\etaSUBATOpSigmaA{\biggerbdeta_{\mbox{\scriptsize$\bA\rightarrow \pDens(\bSigma|\bA)$}}}
\def\mSUBfTOtheta{\biggerm_{\mbox{\scriptsize$f\rightarrow\theta$}}(\theta)}
\def\mSUBthetaTOf{\biggerm_{\mbox{\scriptsize$\theta\rightarrow f$}}(\theta)}
\def\etaSUBfTOtheta{\biggerbdeta_{\mbox{\scriptsize$f\rightarrow\theta$}}}
\def\etaSUBfCONNtheta{\biggerbdeta_{\mbox{\scriptsize$f\leftrightarrow\theta$}}}
\def\etaSUBthetaTOf{\biggerbdeta_{\mbox{\scriptsize$\theta\rightarrow f$}}}
\def\mSUBpSigmaATOSigma{\biggerm_{\mbox{\scriptsize$\pDens(\bSigma|\bA)\rightarrow\bSigma$}}(\bSigma)}
\def\mSUBpSigmaATOA{\biggerm_{\mbox{\scriptsize$\pDens(\bSigma|\bA)\rightarrow\bA$}}(\bA)}
\def\mSUBSigmaTOpSigmaA{\biggerm_{\mbox{\scriptsize$\bSigma\rightarrow \pDens(\bSigma|\bA)$}}(\bSigma)}

\def\mSUBATOpSigmaA{\biggerm_{\mbox{\scriptsize$\bA\rightarrow \pDens(\bSigma|\bA)$}}(\bA)}
\def\mSUBpThetaTOTheta{\biggerm_{\mbox{\scriptsize$\pDens(\bTheta)\rightarrow\bTheta$}}(\bTheta)}

\def\etaSUBpthetaTOtheta{\biggerbdeta_{\mbox{\scriptsize$\pDens(\theta)\rightarrow\theta$}}}
\def\etaSUBpThetaTOTheta{\biggerbdeta_{\mbox{\scriptsize$\pDens(\bTheta)\rightarrow\bTheta$}}}
\def\GSUBpThetaTOTheta{G_{\mbox{\scriptsize$\pDens(\bTheta)\rightarrow\bTheta$}}}
\def\etaSUBpSigmaACONNSigma{\biggerbdeta_{\mbox{\scriptsize$\pDens(\bSigma|\bA)\leftrightarrow\bSigma$}}}
\def\etaSUBpSigmaACONNA{\biggerbdeta_{\mbox{\scriptsize$\pDens(\bSigma|\bA)\leftrightarrow\bA$}}}

\def\etaSUBpsigsqaTOsigsq{\biggerbdeta_{\mbox{\scriptsize$\pDens(\sigma^2|a)\rightarrow\sigma^2$}}}
\def\etaSUBpsigsqaTOa{\biggerbdeta_{\mbox{\scriptsize$\pDens(\sigma^2|a)\rightarrow a$}}}
\def\etaSUBsigsqTOpsigsqa{\biggerbdeta_{\mbox{\scriptsize$\sigma^2\rightarrow \pDens(\sigma^2|a)$}}}
\def\etaSUBaTOpsigsqa{\biggerbdeta_{\mbox{\scriptsize$a\rightarrow \pDens(\sigma^2|a)$}}}

\def\GSUBpSigmaATOSigma{G_{\mbox{\scriptsize$\pDens(\bSigma|\bA)\rightarrow\bSigma$}}}

\def\GSUBSigmaTOpSigmaA{G_{\mbox{\scriptsize$\bSigma\rightarrow \pDens(\bSigma|\bA)$}}}
\def\GSUBpSigmaATOA{G_{\mbox{\scriptsize$\pDens(\bSigma|\bA)\rightarrow\bA$}}}
\def\GSUBATOpSigmaATAB{{\small G}_{\mbox{\tiny$\bA\rightarrow \pDens(\bSigma|\bA)$}}}
\def\GSUBATOpSigmaA{G_{\mbox{\scriptsize$\bA\rightarrow \pDens(\bSigma|\bA)$}}}
\def\GSUBaTOpsigsqa{G_{\mbox{\scriptsize$a\rightarrow \pDens(\sigma^2|a)$}}}
\def\GSUBsigsqTOpsigsqa{G_{\mbox{\scriptsize$\sigma^2\rightarrow \pDens(\sigma^2|a)$}}}

\def\etaSUBpbetauSigmaTObetau
{\biggerbdeta_{\mbox{\scriptsize$\pDens(\bbeta,\bu|\bSigma)\rightarrow(\bbeta,\bu)$}}}
\def\etaSUBpbetauSigmaTOSigma
{\biggerbdeta_{\mbox{\scriptsize$\pDens(\bbeta,\bu|\bSigma)\rightarrow\bSigma$}}}
\def\etaSUBbetauTOpbetauSigma
{\biggerbdeta_{\mbox{\scriptsize$(\bbeta,\bu)\rightarrow \pDens(\bbeta,\bu|\bSigma)$}}}
\def\etaSUBSigmaTOpbetauSigma
{\biggerbdeta_{\mbox{\scriptsize$\bSigma\rightarrow \pDens(\bbeta,\bu|\bSigma)$}}}
\def\Gdiag{G_{\mbox{\tiny{\rm diag}}}}
\def\Gfull{G_{\mbox{\tiny{\rm full}}}}
\def\Mone{
\begin{tiny}
\left[ 
\arraycolsep=2.2pt\def\arraystretch{2.2}
\begin{array}{c | c | c | c | c}
\setstretch{5.5}
   \bigtimes & \bigtimes & \bigtimes & \bigtimes & \bigtimes\\
   \hline
   \bigtimes & \bigtimes & \bigtimes & \bigtimes & \bigtimes\\
   \hline
   \bigtimes & \bigtimes & \bigtimes & \bigtimes& \bigtimes\\
   \hline
   \bigtimes & \bigtimes & \bigtimes & \bigtimes& \bigtimes\\
   \hline
   \bigtimes & \bigtimes & \bigtimes & \bigtimes& \bigtimes\\
\end{array}\right]
\end{tiny}
}
\def\Mtwo{
\begin{tiny}
\left[ 
\arraycolsep=2.2pt\def\arraystretch{2.2}
\begin{array}{c | c | c | c | c}
\setstretch{5.5}
   \bigtimes & \bigtimes & \padzero & \padzero & \bigtimes\\
   \hline
   \bigtimes & \bigtimes & \bigtimes & \padzero & \padzero\\
   \hline
   \padzero & \bigtimes & \bigtimes & \padzero & \bigtimes\\
   \hline
   \padzero & \padzero & \padzero & \bigtimes& \padzero\\
   \hline
   \bigtimes & \padzero & \bigtimes & \padzero& \bigtimes\\
\end{array}\right]
\end{tiny}
}
\def\Mthree{
\begin{tiny}
\left[ 
\arraycolsep=2.2pt\def\arraystretch{2.2}
\begin{array}{c | c | c | c | c}
\setstretch{5.5}
   \bigtimes & \padzero & \padzero & \padzero & \bigtimes\\
   \hline
   \padzero & \bigtimes & \bigtimes & \padzero & \padzero\\
   \hline
   \padzero & \bigtimes & \bigtimes & \padzero& \padzero\\
   \hline
   \padzero & \padzero & \padzero & \bigtimes& \padzero\\
   \hline
   \bigtimes & \padzero & \padzero & \padzero& \bigtimes\\
\end{array}\right]
\end{tiny}
}
\def\Mfour{
\begin{tiny}
\left[ 
\arraycolsep=2.2pt\def\arraystretch{2.2}
\begin{array}{c | c | c | c | c}
\setstretch{5.5}
   \bigtimes & \padzero & \padzero & \padzero & \padzero\\
   \hline
   \padzero & \bigtimes & \padzero & \padzero & \padzero\\
   \hline
   \padzero & \padzero & \bigtimes & \padzero& \padzero\\
   \hline
   \padzero & \padzero & \padzero & \bigtimes& \padzero\\
   \hline
   \padzero & \padzero & \padzero & \padzero& \bigtimes\\
\end{array}\right]
\end{tiny}
}